\documentclass[letterpaper, 10 pt, conference]{ieeeconf}  % Comment this line out if you need a4paper

\IEEEoverridecommandlockouts                              % This command is only needed if 
\usepackage{cite}
\usepackage{amsmath,amssymb,amsfonts}
\usepackage{algorithmic}
\usepackage{algorithm}
\usepackage{graphicx}
\usepackage[export]{adjustbox} 
\usepackage{textcomp}
\usepackage{xcolor}
\usepackage[acronyms]{glossaries}
\usepackage{mathtools}
\usepackage[colorinlistoftodos,prependcaption]{todonotes}
\usepackage{regexpatch}
\usepackage{svg}
\usepackage{enumerate}
\usepackage{tablefootnote}
\usepackage{threeparttable}
\makeatletter
\xpatchcmd{\@todo}{\setkeys{todonotes}{#1}}{\setkeys{todonotes}{inline,#1}}{}{}
\makeatother

\usepackage{booktabs}
\usepackage{flushend} % balance last page
\usepackage{caption}
\usepackage{subcaption}
\usepackage{comment}
\usepackage{jabbrv}
\usepackage{multirow}
\usepackage{tikz}
\usepackage{tikz-network}
\usepackage{tikz-3dplot}
\usepackage{pgfmath}
\usetikzlibrary{3d, calc}
\usepackage{blochsphere}
\usepackage{pgfplots}
\pgfplotsset{compat=1.9}
\usetikzlibrary{arrows.meta}
\tikzset{>=Latex}
\usetikzlibrary{shapes.misc}
\tikzset{cross/.style={cross out, draw=black, minimum size=2*(#1-\pgflinewidth), inner sep=0pt, outer sep=0pt},
cross/.default={1pt}}
\usepackage{array} 
\usepackage{dsfont} % this is for the indicator function
\usepackage{hyperref} % links

\def\BibTeX{{\rm B\kern-.05em{\sc i\kern-.025em b}\kern-.08em
    T\kern-.1667em\lower.7ex\hbox{E}\kern-.125emX}}

\DeclareMathOperator*{\argmin}{argmin}
\def\ie{\textit{i.e.},}
\def\eg{\textit{e.g.},}
\def\etc{\textit{etc.},}

\newcommand{\qt}[1]{``#1''}

\def\labelset{L}

\def\figvspace{}
\def\tabcapspace{}

\newacronym{vlm}{VLM}{Visual-Language Model}
\newacronym{llm}{LLM}{Large Language Model}
\newacronym{iou}{IoU}{Intersection over Union}
\newacronym{ap}{AP}{Average Precision}
\newacronym{rbf}{RBF}{Radial Basis Function}
\newacronym{svm}{SVM}{Support Vector Machine}
\newacronym{csvm}{C-SVM}{SVM with cosine distance metric}
\newacronym{knn}{KNN}{K-Nearest Neighbors}
\newacronym{lora}{LoRA}{Low-Rank Adaptation of Large Language Models}

\newacronym{quash}{QuASH}{Query Augmentation using Semantic Heuristics}
\definecolor{figyellow}{HTML}{FDE724}
\definecolor{figpurple}{HTML}{440154}

\title{\LARGE \bf
QuASH: Using Natural-Language Heuristics to \\ Query Visual-Language Robotic Maps %\\ 
}

\author{Matti Pekkanen, Francesco Verdoja, and Ville Kyrki % <-this % stops a space
\thanks{This work was supported by the Research Council of Finland (decision 354909). The authors acknowledge the use of the MIDAS infrastructure of the Aalto School of Electrical Engineering.}
\thanks{M.\ Pekkanen, F.\ Verdoja, and V.\ Kyrki are with the School of Electrical Engineering, Aalto University, Espoo, Finland. {\tt\small \{firstname.lastname\}@aalto.fi}}%
}

\makeatletter
\newcommand\footnoteref[1]{\protected@xdef\@thefnmark{\ref{#1}}\@footnotemark}
\makeatother

\begin{document}

\maketitle
\thispagestyle{empty}
\pagestyle{empty}

\bstctlcite{IEEEexample:BSTcontrol}

%%%%%%%%%%%%%%%%%%%%%%%%%%%%%%%%%%%%%%%%%%%%%%%%%%%%%%%%%%%%%%%%%%%%%%%%%%%%%%%%
\begin{abstract}

Embeddings from Visual-Language Models are increasingly utilized to represent semantics in robotic maps, offering an open-vocabulary scene understanding that surpasses traditional, limited labels. Embeddings enable on-demand querying by comparing embedded user text prompts to map embeddings via a similarity metric. The key challenge in performing the task indicated in a query is that the robot must determine the parts of the environment relevant to the query.

This paper proposes a solution to this challenge. We leverage natural-language synonyms and antonyms associated with the query within the embedding space, applying heuristics to estimate the language space relevant to the query, and use that to train a classifier to partition the environment into matches and non-matches.  
We evaluate our method through extensive experiments, querying both maps and standard image benchmarks. The results demonstrate increased queryability of maps and images. Our querying technique is agnostic to the representation and encoder used, and requires limited training.
\end{abstract}
%%%%%%%%%%%%%%%%%%%%%%%%%%%%%%%%%%%%%%%%%%%%%%%%%%%%%%%%%%%%%%%%%%%%%%%%%%%%%%%%

\section{Introduction}
\label{sec:intro}

Robots must understand the semantic and geometric properties of their environment to perform complex tasks. While the ability of maps to capture semantics has increased with developments in computer vision, most semantic segmentation methods are trained to segment measurements into specified categories. 

\glspl{vlm} are networks that jointly train a visual and language encoder to learn a mapping from text and image inputs into a common visual-language latent space. Modern, efficient transformer architectures \cite{vaswani2017attention}, such as CLIP \cite{radford_learning_2021}, are trained with large data sets, which promises to enable these models to effectively have an \textit{open vocabulary}, meaning visual semantics can be matched to natural language sentences instead of manually chosen categories, allowing richer representations of semantics and helping to overcome the difficulty of adapting to new environments.

\begin{figure}[t]
    \centering
    \includegraphics[width=0.475\textwidth]{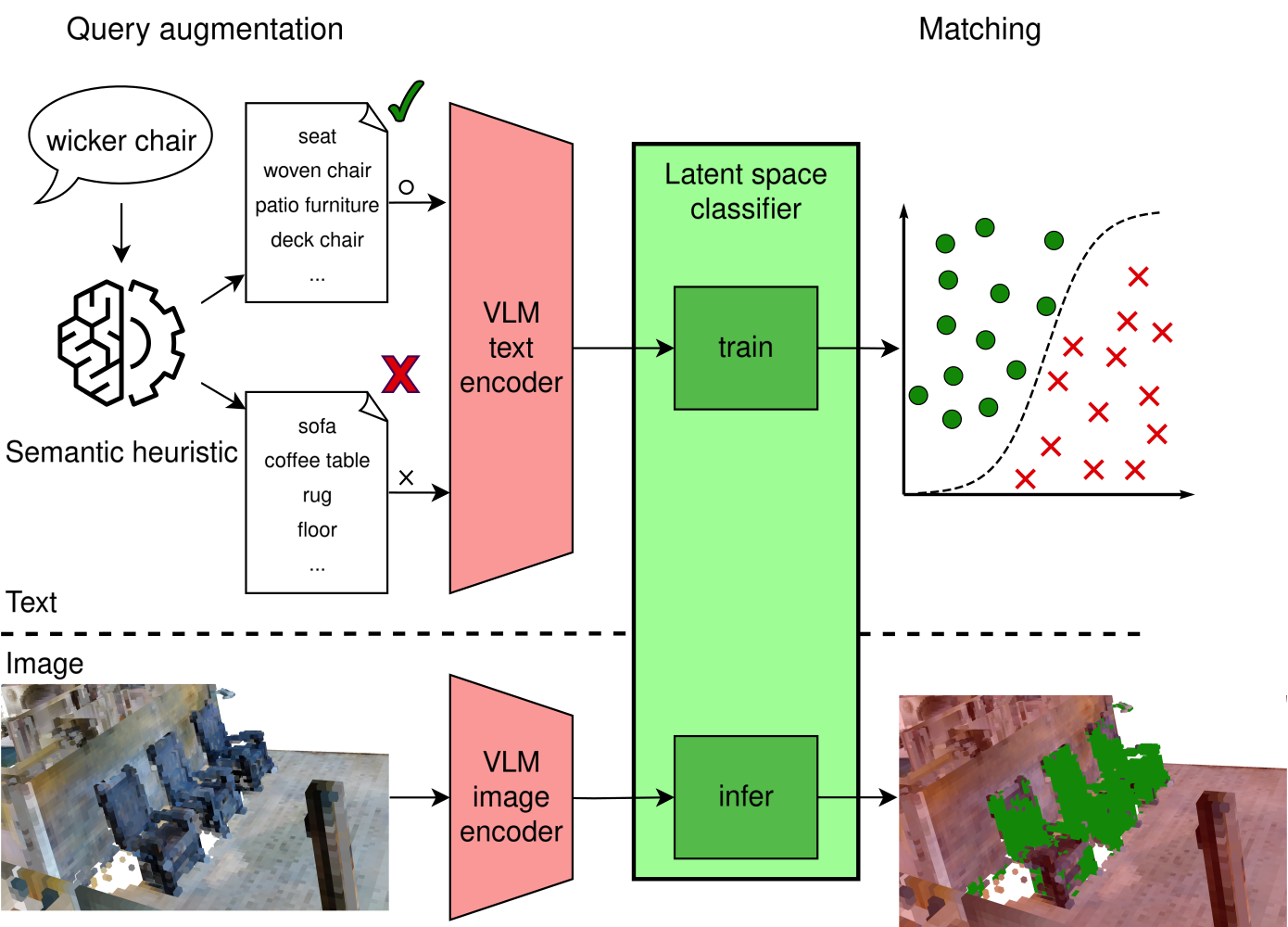}
    \caption{ 
    Our method, QuASH, works by augmenting a single query with a semantic heuristic to acquire a set of semantic synonyms and antonyms to train a classifier in the latent space of a \gls{vlm}. The classifier enables non-linear classification in the latent space, which in turn enables accurate estimation of the region in an image or a map that corresponds to the query.
    }
    \label{fig:header-img}
    \figvspace{}
\end{figure}

Lately, many types of maps \cite{kerr_lerf_2023, huang_visual_2023, yu_language-embedded_2024, jatavallabhula_conceptfusion_2023, gu_conceptgraphs_nodate} using the embeddings of \glspl{vlm}, \ie{} \emph{latent-semantic maps} \cite{pekkanen_2025_latent_semantics}, have been proposed. The idea is that the visual perceptions of the robot are embedded with the visual encoder of a \gls{vlm}, and these latent representations are stored as semantics in the map. These embeddings are then queried on demand by supplying the user query as text, embedding it, and comparing the query embedding to the embeddings in the map using a similarity metric, typically cosine similarity. Despite these advances, the key challenge remains: the selection of relevant matches.

Many methods have been proposed for matching a query to a latent-semantic map, \eg{} by contrasting the query embedding with embeddings of a negative query \cite{huang_visual_2023}, or by using a threshold on a similarity metric \cite{jatavallabhula_conceptfusion_2023}. While methods exist that can utilize direct similarities \cite{qiu_learning_nodate}, most methods ultimately require a decision threshold: some parts of the map are considered matches, while others are not. However, the matching performance of the state-of-the-art is still relatively low, making this an open problem.

In this work, we propose a solution to this problem. We begin by formalizing the problem in terms of identifying and matching sets between the visual, language, and latent spaces. This allows us to formalize the process of querying as the process of classifying the latent-space embeddings. Furthermore, we show how the existing state-of-the-art solutions fit into our formalization.

We propose \emph{\gls{quash}}, a method to query the embeddings, shown in Figure \ref{fig:header-img}, by augmenting the query with a set of semantically close words, as well as producing a set of counterexamples, then training an off-the-shelf classifier on these sets. Our method is not dependent on explicit thresholds, does not require expensive neural network training, and works with any encoder. We show that the proposed method outperforms the comparison with a single complementary query in both standard image benchmarks and in maps. Finally, we conduct extensive ablation studies of querying latent-semantic maps.

The main contributions of this paper are:
\begin{enumerate}[i)]
    \item A formalization of the problem of matching language queries to visual objects, which generalizes most common approaches in the state-of-the-art.
    \item \gls{quash}: a novel encoder-agnostic method\footnote{\href{https://aalto-intelligent-robotics.github.io/quash/}{https://aalto-intelligent-robotics.github.io/quash/}} for querying \gls{vlm}-embeddings. The method is evaluated in the context of querying maps and images.
    \item Extensive ablation experiments on querying \gls{vlm}-embeddings.
\end{enumerate}

\section{Related work}
\label{sec:related}

In this work, we are concerned with the \emph{queryability} of embeddings \cite{pekkanen_2025_latent_semantics}, which differs from two closely related tasks: open-vocabulary object detection and semantic segmentation. 

Most state-of-the-art methods implicitly estimate the region in the embedding space corresponding to the query and use one of two approaches to find matches. First, thresholding the cosine similarities \cite{patel_razer_2025, jatavallabhula_conceptfusion_2023} produces effectively a conic volume centered on the query $q$, shown in Figure \ref{fig:method-img}a. This does not account for the variation in the relevance of the dimensions to $q$. 

Furthermore, by using a single query $q$, the method cannot accurately estimate the extent of the region. The embeddings in the embedding space are located on a very narrow cone \cite{liang_mind_2022} on the surface of a hypersphere, which is so narrow that the embeddings are effectively on an ($N-1$)-dimensional hyperplane. This means that the embeddings acquired by thresholding are in an ($N-1$)-dimensional isotropic hypersphere.

Second, by comparing the embeddings to the query and a negative query, such as \qt{other} \cite{huang_visual_2023, kerr_lerf_2023, wei_ovexp_2024, qiu_learning_nodate}. This produces an implicit estimate of the region by linearly splitting the embedding space into two regions by a hyperplane, shown in Figure \ref{fig:method-img}b. 

Unlike this work, the state-of-the-art methods always produce isotropic and connected estimates of the region. Given that the directions have semantic meaning, with most queries, we expect the directions not to be uniformly important with respect to the query. To estimate the shape of the region, in this work, we estimate the relative importance of the dimensions of the embedding space with respect to $q$ from the spread of the sampled embeddings. 

Other proposed ways to solve the thresholding problem are to train an additional neural network to classify the queries \cite{liu_neural_2025, li_semanticsplat_2025, wang_cris_2022, luddecke_image_2022}, use an \gls{llm} \cite{zhang_mag-nav_2025, linok_open-vocabulary_2025, zhong_language-driven_nodate, gu_conceptgraphs_nodate}, or fine-tune the \gls{vlm} to the query \cite{chapman_queryadapter_2025}. 

Alternatively, many methods do not estimate the region at all, but instead use the whole set of similarities \cite{yuan_uni-fusion_2024, yu_language-embedded_2024, yan_dynamic_2024, sharma_semantic_2023, peng_openscene_2023, qiu_learning_nodate, kim_ov-map_nodate}. In this work, we aim to query the maps directly, without having to train a neural network.

% --------- Global Params ----------
\def\Variant{th} % th | ours | other
\def\R{2}      % circle radius
\def\ofs{1}      % offset
\def\axlen{1.2}      % axis length
\def\diamond{1.5pt}      % diamond size
\def\conelen{1.2}      % dashed line length
\def\bluerad{0.6pt}      % blue circle radius
\def\redrad{0.6pt}      % red circle radius
\def\phiA{48}    % start of red segment (deg)
\def\phiB{68}    % end of red segment (deg)
\def\phiArrow{58}% direction for the black arrow (deg)
\def\phiMarkA{52}% "ours/other" marker angles
\def\phiMarkB{66}

% helper: point on circle at angle (deg)
\newcommand{\pt}[1]{({\R*cos(#1)},{\R*sin(#1)})}
\newcommand{\pto}[3]{({#2+\R*cos(#1)},{#3+\R*sin(#1)})}
\newcommand{\ptom}[4]{({#2+#4*\R*cos(#1)},{#3+#4*\R*sin(#1)})}
\newcommand{\ptm}[2]{({#2*\R*cos(#1)},{#2*\R*sin(#1)})}

% --------- Styles ----------
\tikzset{
  ptblue/.style={line width=0pt, fill=blue!70, draw=none},
  ptred/.style={line width=0pt, fill=red!80!orange, draw=none},
  ptblack/.style={line width=0pt, fill=black, draw=none},
  arc/.style={line width=2pt, draw=none},
  axis/.style={thin, black!60},
  cone/.style={dash pattern=on 6pt off 5pt, line width=1.2pt, draw=violet!70!black},
  tangent/.style={dash pattern=on 4pt off 4pt, line width=1pt, draw=black!70},
  emph/.style={draw=red!80!orange, line width=1.2pt},
  ell/.style={draw=red!80, line width=1pt},
  arr/.style={-Latex, line width=1.4pt},
  query/.style={black!70 }
}

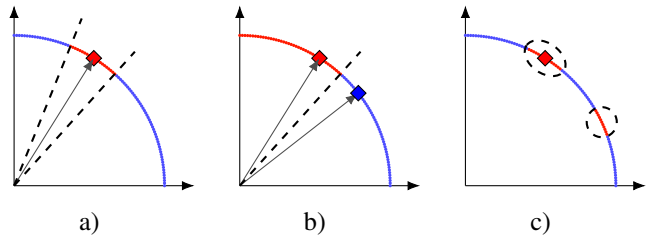
\begin{figure}[t]
\centering
    \begin{tikzpicture}[scale=1]
    %%%%%%%%%%%%%%%%%%%%%%%%%%%%%%%%%%%%%%%%%%%%%%%%%%%%%%%%%%%%%%%%%%%%%%%%%%%%%%%%%%%%%%%%%%%%%%%%%%%%%%%%%%%%%%%%%%
    \coordinate []  (O) at (0,0);
    %\node[align=left] at (0.5*\R, -0.5) {a) Similarity thresholding};
    \node[align=left] at (0.5*\R, -0.5) {a)};
    \draw[->] (O) -- ($(O) + (0, \axlen*\R)$);
    \draw[->] (O) -- ($(O) + (\axlen*\R, 0)$);
    \def\thetaStart{0}
    \def\thetaEnd{90}
    
    % points along the arc
    \foreach \a in {\thetaStart,...,\thetaEnd} {
      \ifthenelse{\lengthtest{\a pt < \phiA pt} \OR \lengthtest{\a pt > \phiB pt}}
        {\fill[ptblue] \pt{\a} circle[radius=\bluerad];}
        {\fill[ptred]  \pt{\a} circle[radius=\redrad];}
    }
    % query
    \filldraw[fill=blue] \pt{\phiArrow} node[diamond,aspect=1,draw,inner sep=\diamond,fill=red]{};
    \draw[query,->] (O) -- \pt{\phiArrow};

    % cone
    \draw[black, dashed, thick] (O) -- \ptm{\phiA}{\conelen};
    \draw[black, dashed, thick] (O) -- \ptm{\phiB}{\conelen};

    %%%%%%%%%%%%%%%%%%%%%%%%%%%%%%%%%%%%%%%%%%%%%%%%%%%%%%%%%%%%%%%%%%%%%%%%%%%%%%%%%%%%%%%%%%%%%%%%%%%%%%%%%%%%%%%%%%
    \coordinate []  (O2) at (\R+\ofs,0);
    \node[align=left] at (\R+\ofs+0.5*\R, -0.5) {b)};
    \draw[->] (O2) -- ($(O2) + (0, \axlen*\R)$);
    \draw[->] (O2) -- ($(O2) + (\axlen*\R, 0)$);

    % points along the arc
    \foreach \a in {\thetaStart,...,\thetaEnd} {
      \ifthenelse{\lengthtest{\a pt < 48 pt}}
        {\fill[ptblue] \pto{\a}{\R+\ofs}{0} circle[radius=\bluerad];}
        {\fill[ptred]  \pto{\a}{\R+\ofs}{0} circle[radius=\redrad];}
    }
    % query
    \filldraw[fill=blue] \pto{\phiArrow}{\R+\ofs}{0} node[diamond,aspect=1,draw,inner sep=\diamond,fill=red]{};
    \draw[query,->] (O2) -- \pto{\phiArrow}{\R+\ofs}{0};
    \draw[query,->] (O2) -- \pto{38}{\R+\ofs}{0};

    % complement
    \filldraw[fill=blue] \pto{38}{\R+\ofs}{0} node[diamond,aspect=1,draw,inner sep=\diamond,fill=blue]{};

    \draw[black, dashed, thick] (O2) -- \ptom{48}{\R+\ofs}{0}{\conelen};

    %%%%%%%%%%%%%%%%%%%%%%%%%%%%%%%%%%%%%%%%%%%%%%%%%%%%%%%%%%%%%%%%%%%%%%%%%%%%%%%%%%%%%%%%%%%%%%%%%%%%%%%%%%%%%%%%%%
    \coordinate []  (O3) at (2*\R+2*\ofs,0);
    \node[align=left] at (2*\R+2*\ofs+0.5*\R, -0.5) {c)};
    \draw[->] (O3) -- ($(O3) + (0, \axlen*\R)$);
    \draw[->] (O3) -- ($(O3) + (\axlen*\R, 0)$);

    % points along the arc
    \foreach \a in {\thetaStart,...,\thetaEnd} {
        \pgfmathparse{ ((\a<66) && (\a>50)) || ((\a<31) && (\a>19)) }
        \ifnum\pgfmathresult>0
          {\fill[ptred]  \pto{\a}{2*\R+2*\ofs}{0} circle[radius=\redrad];}
        \else
          {\fill[ptblue] \pto{\a}{2*\R+2*\ofs}{0} circle[radius=\bluerad];}
        \fi
    }

    % query
    \filldraw[fill=blue] \pto{\phiArrow}{2*\R+2*\ofs}{0} node[diamond,aspect=1,draw,inner sep=\diamond,fill=red]{};
    
    \draw[black, dashed, thick, rotate around={145:\pto{\phiArrow}{2*\R+2*\ofs}{0}}]
        \pto{\phiArrow}{2*\R+2*\ofs}{0} ellipse [x radius=0.3, y radius=0.2];

    \draw[black, dashed, thick, rotate around={115:\pto{25}{2*\R+2*\ofs}{0}}]
        \pto{25}{2*\R+2*\ofs}{0} ellipse [x radius=0.2, y radius=0.2];
    %%%%%%%%%%%%%%%%%%%%%%%%%%%%%%%%%%%%%%%%%%%%%%%%%%%%%%%%%%%%%%%%%%%%%%%%%%%%%%%%%%%%%%%%%%%%%%%%%%%%%%%%%%%%%%%%%%
    \end{tikzpicture}   
    \caption{ 
        The state-of-the-art solutions implicitly classify the embedding space linearly by thresholding cosine similarities (a) \cite{jatavallabhula_conceptfusion_2023}, or by using a single complementary embedding (b) \cite{huang_visual_2023}. By using a \gls{svm} with a non-linear kernel, our proposed method (c) can classify the embeddings non-linearly. Embeddings matching the query are visualized in red, and those not matching are visualized in blue. A diamond represents the query and its complement, and dashed lines represent the thresholds.
    }
    \label{fig:method-img}
    \figvspace{}
\end{figure}
\section{Problem statement}
\label{sec:problem}

We consider the problem of determining a target object referred to by the query $q$ defined in a language space $\mathcal{L}$. The search space is the internal world representation of the robot, which we denote as the visual space $\mathcal{V}$, \eg{} a map of the environment or an image from the camera. Elements $v \in \mathcal{V}$ are either the pixels or voxels representing a portion of the space. We assume $q$ to correspond to a subset $\mathcal{V}_{\mathcal{Q}} \subset \mathcal{V}$ representing the extent of the target object. The problem is, therefore, to find the mapping $f_q: q \mapsto \mathcal{V}_{\mathcal{Q}}$. 

The underlying assumption behind the existance of $f_q$ shares some commonalities to that behind \glspl{vlm}, which are designed to embed elements of $\mathcal{L}$ and $\mathcal{V}$ into the shared embedding space $\mathcal{E}$ using a pair of embedding functions $\varphi: \mathcal{L} \to \mathcal{E}$ and $\varrho: \mathcal{V} \to \mathcal{E}$ such that for a textual description $l \in \mathcal{L}$ and a corresponding visual point $v \in \mathcal{V}$, $\varphi(l) = \varrho(v)$. 

Therefore, given the \gls{vlm} $M = (\varphi, \varrho)$, our problem corresponds to the problem of finding the set $\mathcal{R_{\mathcal{Q}}} \subset \mathcal{E} = \{ \forall v \in \mathcal{V}_\mathcal{Q} \, : \, \varrho(v)\}$ that contains all embeddings matching the object described by the query $q$. 
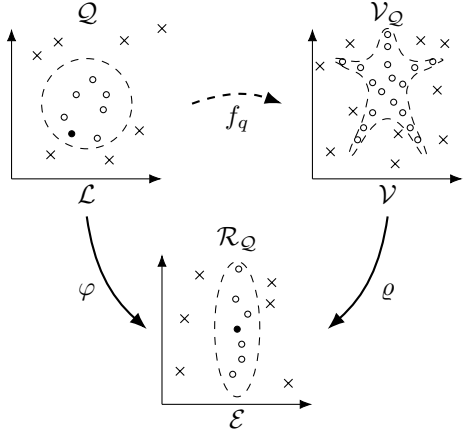
\begin{figure}[t]
\centering
    \begin{tikzpicture}[scale=2]
    
        % Axes
        % language
        \draw[->] (0, 0) -- (0, 1);
        \draw[->] (0, 0) -- (1, 0);

        % visual
        \draw[->] (2, 0) -- (2, 1);
        \draw[->] (2, 0) -- (3, 0);

        % embedding
        \draw[->] (1, -1.5) -- (1, -0.5);
        \draw[->] (1, -1.5) -- (2, -1.5);
    
        % Labels
        \node at (0.5, -0.1) {$\mathcal{L}$};
        \node at (2.5, -0.1) {$\mathcal{V}$};
        \node at (1.5, -1.6) {$\mathcal{E}$};
        \node at (0.5, 1.1) {$\mathcal{Q}$};
        \node at (1.5, -0.4) {$\mathcal{R_{\mathcal{Q}}}$};
        \node at (2.5, 1.1) {$\mathcal{V}_{\mathcal{Q}}$};
    
        % Dashed language domain circle
        \draw[dashed] (0.5, 0.5) circle (0.3);
    
        % Dashed embeddings codomain ellipse
        \draw[dashed] (1.5, -1) ellipse (0.15 and 0.45);

        % Mapping arrows
        \draw[->, thick] (0.5, -0.25) to[bend right=20] (0.9, -1);
        %\draw[->, thick] (2.1, -1) to[bend right=20] (2.5, -0.25);
        \draw[->, thick] (2.5, -0.25) to[bend left=20] (2.1, -1);
        \draw[->, thick, dashed] (1.2, 0.5) to[bend left=20] (1.8, 0.5);
        \node at (1.5, 0.4) {$f_q$};
        \node at (0.5, -0.75) {$\varphi$};
        \node at (2.5, -0.75) {$\varrho$};
    
        % Language Domain points
        \foreach \x/\y in {0.35/0.41, 0.57/0.27, 0.61/0.46, 0.43/0.56, 0.63/0.56, 0.54/0.66} {
            \filldraw[fill=white] (\x,\y) circle (0.02);
        }

        \foreach \x/\y in {0.75/0.92, 0.26/0.16, 0.65/0.125, 0.85/0.32, 0.17/0.82, 0.31/0.91} {
            \draw (\x,\y) node[cross=2pt,rotate=0,black] {};
        }
          
        \filldraw[fill=black] (0.4,0.3) circle (0.02);
        
        % Embedding domain points
        \foreach \x/\y in {1.47/-1.3, 1.51/-0.6, 1.49/-0.8, 1.53/-1.2, 1.57/-0.9, 1.53/-1.1} {
            \filldraw[fill=white] (\x,\y) circle (0.02);
        }
        \foreach \x/\y in {1.12/-1.28, 1.25/-0.64, 1.15/-0.93, 1.72/-0.83, 1.84/-1.36, 1.73/-0.69} {
            \draw (\x,\y) node[cross=2pt,rotate=0,black] {};
        }
        \filldraw[fill=black] (1.5,-1) circle (0.02);
      
        % new stick man figure
        \foreach \x/\y in {
          % legs
          2.30/0.26, 2.70/0.26,
          2.34/0.34, 2.65/0.34,
          2.40/0.44, 2.60/0.44,     
          % torso
          2.46/0.58, 2.54/0.51,
          2.40/0.64, 2.60/0.63,
          2.50/0.76,    
          % shoulders / arms
          2.32/0.74, 2.68/0.74,
          2.20/0.78, 2.80/0.78,   
          % head
          2.50/0.88, 2.50/0.96
        }{
          \filldraw[fill=white] (\x,\y) circle (0.02);
        }
        
        % Crosses (outside the figure)
        \foreach \x/\y in {
          2.05/0.75, 
          2.27/0.45, 
          2.12/0.18, 
          2.57/0.30, 
          2.55/0.10,
          2.25/0.90, 
          2.72/0.92, 
          2.83/0.59,
          2.87/0.26, 
          2.88/0.90, 
        }{
          \draw (\x,\y) node[cross=2pt,rotate=0,black] {};
        }
        
        % Emphasized black dot
        \filldraw[fill=white] (2.53,0.67) circle (0.02);
        
        % --- Dashed contour around the figure ---
        % Tweak 'tension' and coordinates to tighten/loosen the wrap.
        \draw[black, dashed, line cap=round]
          plot[smooth cycle, tension=0.9] coordinates {
            (2.35,0.65)  % left torso
            (2.20,0.75)  % left mid-arm
            (2.15,0.80)  % left hand region
            (2.37,0.80)  % left shoulder
            (2.50,1.00)  % head top
            (2.63,0.80)  % right shoulder
            (2.85,0.80)  % right hand region
            (2.80,0.75)  % right mid-arm
            (2.65,0.65)  % right torso
            (2.65,0.45)  % right hip
            (2.75,0.15)  % right foot
            (2.50,0.40)  % between feet
            (2.25,0.15)  % left foot
            (2.35,0.45)  % left hip
          };
  
    \end{tikzpicture}    
    \caption{ 
        The intuition behind our method is that, as the textual descriptions and visual features are embedded into the same space $\mathcal{E}$, training a classifier on the embeddings of the text descriptions will result in a good classifier in the visual domain. Instead of using just a single example, \ie{} the query $q$, represented as a black dot, we additionally sample semantically close words, represented as circles, to form a set $\mathcal{Q}$, as well as a non-matching queries $\bar{q} \in \bar{\mathcal{Q}}$, represented as crosses, from the language space $\mathcal{L}$. Using these, we aim to capture the extent of the region $\mathcal{R_{\mathcal{Q}}}$ in the embedding space $\mathcal{E}$. Knowing $\mathcal{R_{\mathcal{Q}}}$ enables finding the visual region $\mathcal{V}_{\mathcal{Q}}$ corresponding to the query $q$. 
    }
    \label{fig:intuition}
    \figvspace{}
\end{figure}
\section{Methods}
\label{sec:methods}

To solve this problem, we present \gls{quash}. First, we formalize the process of querying latent-semantic maps in Section \ref{sec:methods-preliminaries}. Second, we present state-of-the-art solutions within our formalization in Section \ref{sec:methods-method-sota}, and finally, propose a querying method by augmenting the query using semantic heuristics in Section \ref{sec:methods-method-ours}.

%
%
%
%%%%%%%%%%%%%%%%%%%%%%%%%%%%%%%%%%%%%%%%%%%%%%%
\subsection{Formalization of the querying}
\label{sec:methods-preliminaries}

Let $\mathcal{Q} \subset \mathcal{L}$ be the preimage of $\mathcal{R}_\mathcal{Q}$ under $\varphi$ and $\mathcal{\bar{Q}} \subset \mathcal{L}$ be the preimage of $\mathcal{E}\setminus\mathcal{R}_\mathcal{Q}$ under $\varphi$. We assume $q \in \mathcal{Q}$. The set $\mathcal{Q}$ contains the set of all semantically close words to $q$. To preserve intuition, we hereafter refer to them as \textit{semantic synonyms}, notwithstanding that they are not literal synonyms. $\mathcal{\bar{Q}}$, on the other hand, contains all negative queries of the query $q$, which we will refer to as \textit{semantic antonyms} for the aforementioned reason.

The common approach in literature is to estimate $\mathcal{R}_{\mathcal{Q}}$ using some classification function $\psi_{\mathcal{Q}}: \mathcal{E} \to \{0, 1\}, \psi_{\mathcal{Q}} = \mathds{1}_{\mathcal{R}_{\mathcal{Q}}}$. To do this, most approaches rely on the implicit assumption that $\psi_{\mathcal{Q}}$ can be learned by embedding samples from $\mathcal{Q}$ and $\mathcal{\bar{Q}}$ using $\varphi$ \cite{huang_visual_2023, kerr_lerf_2023, wei_ovexp_2024, qiu_learning_nodate, jatavallabhula_conceptfusion_2023, patel_razer_2025}. 

More specifically, given a map or image $\mathcal{V}_m \subset \mathcal{V}$, with corresponding embeddings $\mathcal{E}_m \subset \mathcal{E}$, the embedding of the query $e_q = \varphi(q)$ is used as representative sample from $\mathcal{R}_{\mathcal{Q}}$, and an the embedding $e_{\bar{q}} = \varphi(\bar{q})$ of an arbitrary antonym $\bar{q} \in \bar{\mathcal{Q}}$ (\eg{} the word "other") is used as representative sample of $\mathcal{E}\setminus\mathcal{R}_\mathcal{Q}$. 

%
%
%
%%%%%%%%%%%%%%%%%%%%%%%%%%%%%%%%%%%%%%%%%%%%%%%
\subsection{Querying latent-semantic maps: the state of the art}
\label{sec:methods-method-sota}
The classifier $\psi_{\mathcal{Q}}$ is then often either $d_{cos}(\mathcal{E}_m, e_q) < \theta$ \cite{jatavallabhula_conceptfusion_2023, patel_razer_2025}, where cosine distance $d_{cos}$ is thresholded with $\theta$, yielding a hyperconic section of $\mathcal{E}$, shown in Figure \ref{fig:method-img}a, or
\begin{equation}
    \mathds{1}_{e_q} \left( \argmin_{x \in \{e_q, e_{\bar{q}}\}} d_{cos}(\mathcal{E}_m, x) \right),     
\end{equation}
where the embeddings $\mathcal{E}_m$ with $e_q$ and $e_{\bar{q}}$ are compared, splitting $\mathcal{E}$ with a hyperplane \cite{huang_visual_2023, kerr_lerf_2023, wei_ovexp_2024, qiu_learning_nodate}, shown in Figure \ref{fig:method-img}b.

Provided that the directions in the embedding space have semantic meaning, we expect that the directions of the embedding space have varying degrees of relevance with respect to $q$, which means that the subset $\mathcal{R_{\mathcal{Q}}}$ is neither necessarily isotropic nor a connected space, reducing the effectiveness of distance-based classifiers. We argue that estimating the extent of $\mathcal{R_{\mathcal{Q}}}$ from a pair of points $e_q$ and $e_{\bar{q}}$ is not sufficient. 

%
%
%
%%%%%%%%%%%%%%%%%%%%%%%%%%%%%%%%%%%%%%%%%%%%%%%
\subsection{Querying latent-semantic maps: QuASH}
\label{sec:methods-method-ours}
In practice, it is not feasible to acquire $\mathcal{Q}$ or $\bar{\mathcal{Q}}$. We assume that $\mathcal{Q}$ contains semantically similar phrases to the query. This means that they are \qt{close} in natural language, and therefore, the intuition behind our method, shown in Figure \ref{fig:intuition}, is to use a natural-language semantic heuristic to estimate the augmented query set $\mathcal{Q}$.

Sampling from $\mathcal{Q}$ we acquire the semantic synonyms $\mathcal{S}_q$ and sampling $\bar{\mathcal{Q}}$ the semantic antonyms $\mathcal{A}_q$. The embedded samples of synonyms and antonyms form the training data set, with which a classifier $\psi_{\mathcal{Q}}$ is learned in the latent space $\mathcal{E}$. The classifier enables the estimation of the volume of $\mathcal{R}_{\mathcal{Q}} \in \mathcal{E}$. If a non-linear classifier is used, a non-linear estimate of $\mathcal{R}_{\mathcal{Q}}$ can be acquired, as shown in Figure \ref{fig:method-img}c.

The method has two key design choices. First, finding a suitable heuristic function $h(q) = ( \mathcal{S}_q, \mathcal{A}_q )$, as the performance of the classifier depends on the quality of the samples. Second, finding a suitable classifier $\psi_{\mathcal{Q}}$ that can learn the region $\mathcal{R}_{\mathcal{Q}}$ effectively from the samples. We discuss the choices for $h$ in Section \ref{sec:exp-ablation-synonyms}, and for $\psi_{\mathcal{Q}}$ in Section \ref{sec:exp-querying}.

Our method works with the following steps:
\begin{enumerate}
    \item Given the textual query $q$, we create a set of $N$ strings $\mathcal{S}_q \sim \mathcal{Q}$, semantic synonyms to $q$, and $M$ strings $\mathcal{A}_q  \sim \bar{\mathcal{Q}}$ that are semantic antonyms to $q$. Additionally, we add to $\mathcal{A}_q$ a set of generic negative queries found in the literature: \qt{other}, \qt{background}, \qt{nothing}.
    \item Given an embedding function $\varphi$, we embed all strings $\mathcal{E}_{\mathcal{S}_q} = \varphi(\mathcal{S}_q \cup q)$, and $\mathcal{E}_{\mathcal{A}_q} = \varphi(\mathcal{A}_q \cup \mathcal{C})$.
    \item We train a classifier $\psi_{\mathcal{Q}}$ with positive examples $\mathcal{E}_{\mathcal{S}_q}$ and negative examples $\mathcal{E}_{\mathcal{A}_q}$. 
    \item Given a map $\mathcal{M}$ where each map item $m_i$ contain an embedding $e_i$, we acquire a binary mask over the map $\mathcal{V}_{\mathcal{Q}} = \psi_{\mathcal{Q}}(\mathcal{M})$
\end{enumerate}

The decision boundary of $\psi_{\mathcal{Q}}$ forms the estimate of $\mathcal{R_{\mathcal{Q}}}$.

\section{Experiments}
\label{sec:experiments}

The main question that we aim to answer with the experiments is: Does using the natural-language heuristics improve the visual classification task?

To answer this question, we evaluated our proposed querying method in a series of experiments. We employ the evaluation method described in \cite{pekkanen_2025_latent_semantics} to assess \emph{queryability}.

%
%
%
%
%
%%%%%%%%%%%%%%%%%%%%%%%%%%%%%%%%%%%%%%%%%%%%%%%%%%%%%%%%%%%%%%%%%%%%%%%%%%%%%%%%%%%%%%%%%%%%%%%%%%%%%
\subsection{Baseline}
\label{sec:exp-baseline}

In all experiments, we use as the \emph{baseline} a method proposed in \cite{huang_visual_2023}, where the embedding of each voxel is compared with the embeddings of the query and the complement, the string \qt{other}. 

The baseline uses \emph{prompt engineering} (PE), where both the query and the complement are inserted into a list of 63 strings, such as \qt{There is $q$ in the scene} or \qt{a photo of $q$}. All the resulting strings are embedded, and the means of the embeddings are used as the query and the complement. The original method additionally post-processes the query results and evaluates in 2D, but these steps decrease performance; therefore, our baseline method does not use them.

The embeddings that are closer to the query than to the complement with respect to the cosine similarity metric are chosen as the results. We evaluate the effect of prompt engineering in an ablation study, presented at Section \ref{sec:exp-ablation-prompt}.

%
%
%
%
%
%%%%%%%%%%%%%%%%%%%%%%%%%%%%%%%%%%%%%%%%%%%%%%%%%%%%%%%%%%%%%%%%%%%%%%%%%%%%%%%%%%%%%%%%%%%%%%%%%%%%%
\subsection{Data set and metrics}
\label{sec:exp-data}

In all map querying experiments, 10 maps, the same as in \cite{huang_visual_2023}, were created from the Matterport3D data set \cite{chang_matterport3d_2017}, containing ground truth semantic segmentation of 40 labels $\labelset$, using a 3D version of VLMaps \cite{huang_visual_2023, VLMaps_software}, where the method works similarly, but is extended to 3D. We use the original proposed parametrization. We use two encoders, LSeg \cite{li_language-driven_2022} and OpenSeg \cite{ghiasi_scaling_2022}, with pre-trained weights provided by the original authors.

We removed seven negative or aggregate classes: \qt{void}, \qt{unlabeled}, \qt{misc}, \qt{objects}, \qt{seating}, \qt{furniture}, and \qt{appliances}, from $\labelset$ to acquire query set $\labelset_q$\footnote{\label{note:appendix}\href{https://aalto-intelligent-robotics.github.io/quash/}{https://aalto-intelligent-robotics.github.io/quash/}} of 34 labels. 

In all experiments, following \cite{pekkanen_2025_latent_semantics}, we compute precision, recall, F1-score, and \gls{iou} as metrics and report the average over the data set. As the areas of interest are relatively small in our experiments, true negatives dominate accuracy, making it an unsuitable metric.

%
%
%
%
%
%%%%%%%%%%%%%%%%%%%%%%%%%%%%%%%%%%%%%%%%%%%%%%%%%%%%%%%%%%%%%%%%%%%%%%%%%%%%%%%%%%%%%%%%%%%%%%%%%%%%%
\subsection{Querying method}
\label{sec:exp-querying}

To test our proposed method of querying \gls{vlm} embeddings, we evaluated querying maps, described in Section \ref{sec:exp-maps}, as well as performed a set of standard image benchmarks, described in Section \ref{sec:exp-img}.

In our method, we prompted\footnoteref{note:appendix} GPT-4o to produce a list of 20 words that are: synonyms, closely related words, subclasses, superclasses, related objects, \etc{} of the query to acquire $\mathcal{S}_q$ and similarly 20 antonyms to acquire $\mathcal{A}_q$, which we used as the heuristic for $\mathcal{Q}$. 

As the classifier $\psi$, we evaluated all of the applicable off-the-shelf classifiers from the scikit-learn library, and decided to use \gls{svm} with \gls{rbf} kernel with Euclidean and cosine similarity metrics, as it had the best performance.

\subsubsection{Image benchmarks}
\label{sec:exp-img}

First, we tested our method against the baseline in two widely used image benchmarks: COCO 80 and Pascal Context 459. Additionally, we use the set of images used in the map generation as a data set, denoted \qt{Matterport}. In these experiments, only LSeg was used as the encoder.

\subsubsection{Querying maps}
\label{sec:exp-maps}

Second, we tested our method in querying maps created as described in Section \ref{sec:exp-data}.

We compared our method with the baseline. Additionally, in this experiment, we compare to another method which uses the best information available, which we will refer to as \qt{ground truth}. We use this method to separate the evaluation of the generation of the $\mathcal{Q}$ and its use. The method uses the whole set of 1659 $\labelset_{sup}$ classes present in the Matterport3D data set as the heuristic for $\mathcal{Q}$. $\labelset_{sup}$ contains more specific subclasses of the superclasses that compose the set $\labelset$. For example, $\labelset$ contains the superclass \qt{chair}, and $\labelset_{sup}$ contains the subclasses \qt{armchair}, \qt{dining chair}. So, for each query term in $\labelset$, we selected up to 20 random specific subclasses of that query as synonyms and, as antonyms, 100 random other subclasses that do not correspond to the query.

%
%
%
%
%
%%%%%%%%%%%%%%%%%%%%%%%%%%%%%%%%%%%%%%%%%%%%%%%%%%%%%%%%%%%%%%%%%%%%%%%%%%%%%%%%%%%%%%%%%%%%%%%%%%%%%
\subsection{Ablation studies}
\label{sec:exp-ablation}

\subsubsection{Prompt engineering}
\label{sec:exp-ablation-prompt}

We tested the effect of prompt engineering, as described in Section \ref{sec:exp-baseline}, on the baseline method and our method.

\subsubsection{Antonyms}
\label{sec:exp-ablation-complement}

We evaluated the baseline method with a set of antonyms of the query found in the literature: \qt{other} \cite{li_language-driven_2022, huang_visual_2023, wei_ovexp_2024}, \qt{object} \cite{kerr_lerf_2023}, \qt{things} \cite{kerr_lerf_2023}, \qt{stuff} \cite{kerr_lerf_2023}, \qt{texture} \cite{kerr_lerf_2023}, \qt{background} \cite{miller_open-set_2024}, and a zero vector \cite{miller_open-set_2024}, as well as \qt{nothing} and \qt{others}.

\subsubsection{Language distribution}
\label{sec:exp-ablation-synonyms}

We evaluated four different methods of creating the query sets $\mathcal{Q}$ and $\bar{\mathcal{Q}}$: \gls{llm}, lexicon, adjectives, and ground truth. 

The \qt{\gls{llm}} is the default implementation of our method, as described in Section \ref{sec:exp-querying}. The \qt{ground truth} set is described in Section \ref{sec:exp-maps}. 

In the \qt{lexicon} method, we select 20 random words from the English lexicon for synonyms and 20 random words for antonyms. In the \qt{adjectives} method, we produce a static list of 44 adjectives across seven types, \eg{} different colors, materials, and textures. For each query $q$, we add to $\mathcal{S}_q$ all singular adjectives (\eg{} color + $q$), 10 randomly combined adjectives from two types (\eg{} color + material + $q$), and five randomly combined adjectives from three types, totaling 59 random variations of the query. As antonyms, a similar set is created by using \qt{other} instead of the query.

\section{Results}
\label{sec:results}

%--%--%--%--%--%--%--%--%--%--%--%--%--%--%--%--%--%--%--%--%--%--%--%--%--%--%--%--
\begin{table}[t]
    \caption{Results of image benchmark experiment with LSeg}
    \tabcapspace{}
    \begin{center}
        \setlength\tabcolsep{6pt}
        \scalebox{1.0}{
                \begin{tabular}{llllll}
                    \toprule
                    \textbf{Dataset} & \textbf{Method} & \textbf{F1} & \textbf{IoU} & \textbf{Precision} & \textbf{Recall} \\
                    \midrule
                    \multirow{3}{*}{COCO}       & baseline  & 0.680          & 0.515          & 0.536          & \textbf{0.929} \\
                                                & Our SVM   & \textbf{0.786} & \textbf{0.647} & \textbf{0.696} & 0.901          \\
                                                & Our C-SVM & 0.771          & 0.627          & 0.672          & 0.905          \\
                    \midrule
                    \multirow{3}{*}{PC 459}     & baseline  & 0.654          & 0.486          & 0.540          & \textbf{0.831} \\
                                                & Our SVM   & 0.667          & 0.501          & \textbf{0.693} & 0.643          \\
                                                & Our C-SVM & \textbf{0.728} & \textbf{0.572} & 0.680          & 0.783          \\
                    \midrule
                    \multirow{3}{*}{Matterport} & baseline  & 0.544          & 0.374          & 0.448          & 0.691          \\
                                                & Our SVM   & \textbf{0.621} & \textbf{0.451} & \textbf{0.626} & 0.616          \\
                                                & Our C-SVM & 0.579          & 0.407          & 0.483          & \textbf{0.724} \\
                    \bottomrule
                \end{tabular}
        }
        \setlength\tabcolsep{6pt}
        \label{tab:image-benchmarks}
    \end{center}
    \figvspace{}
\end{table}

%--%--%--%--%--%--%--%--%--%--%--%--%--%--%--%--%--%--%--%--%--%--%--%--%--%--%--%--

\newcommand\hsheep{3cm}
\newcommand\hfridge{3cm}
\newcommand\hperson{3cm}
\newcommand\hsb{3cm}
\newcommand\divCoco{0.25}
\newcommand\heit{4cm}
\newcommand\wis{4cm}

\begin{figure*}[ht!]
    \begin{subfigure}[t]{\divCoco\textwidth}
        \centering
        \includegraphics[height=\hsheep]{746-sheep-COCO_val2014_000000078522-orig.png}
        \caption{Original: Sheep}
        \label{fig:threshold}
    \end{subfigure}%
    \begin{subfigure}[t]{\divCoco\textwidth}
        \centering
        \includegraphics[height=\hfridge]{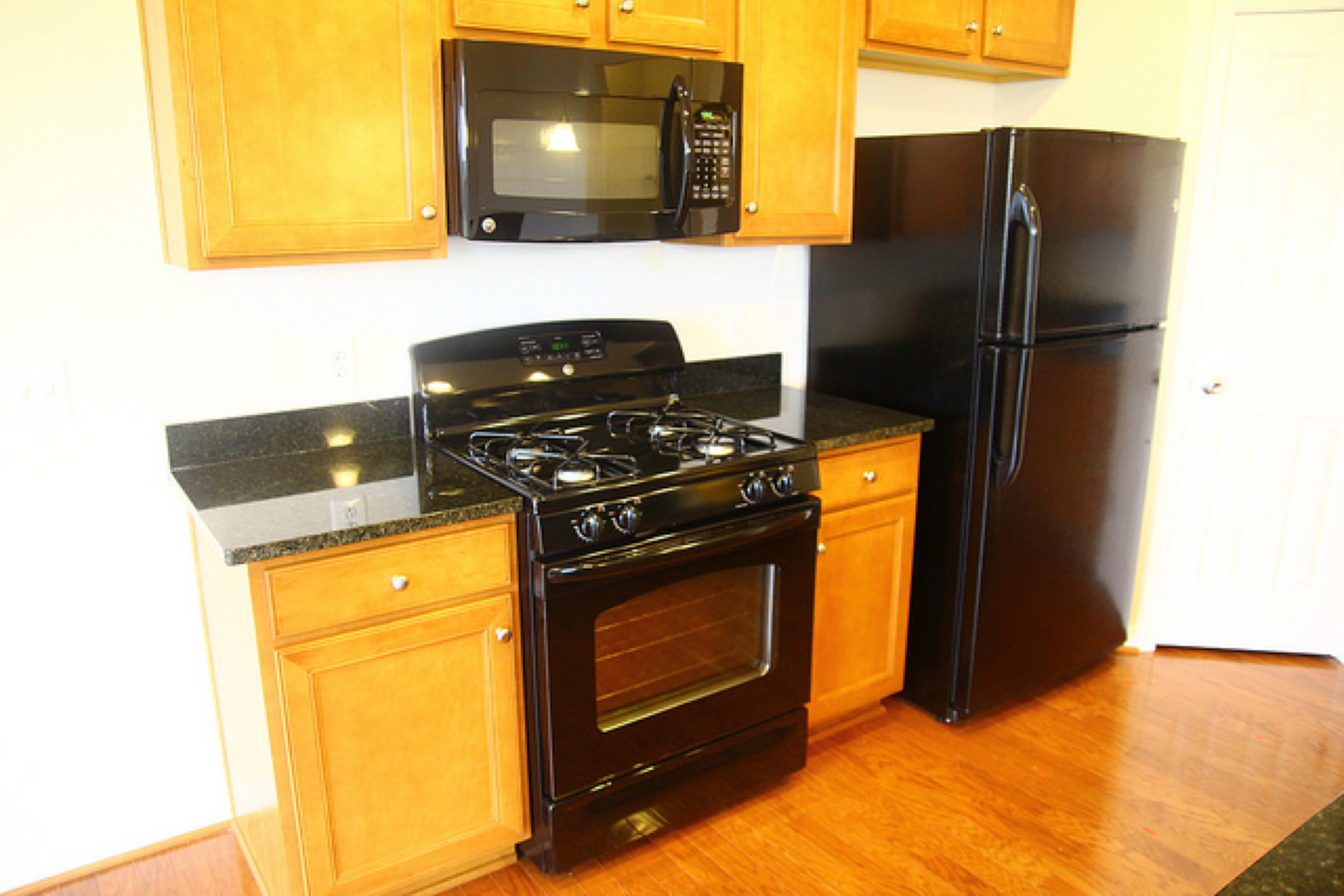}
        \caption{Original: Refrigerator}
        \label{fig:threshold}
    \end{subfigure}%
    \begin{subfigure}[t]{\divCoco\textwidth}
        \centering
        \includegraphics[height=\hperson]{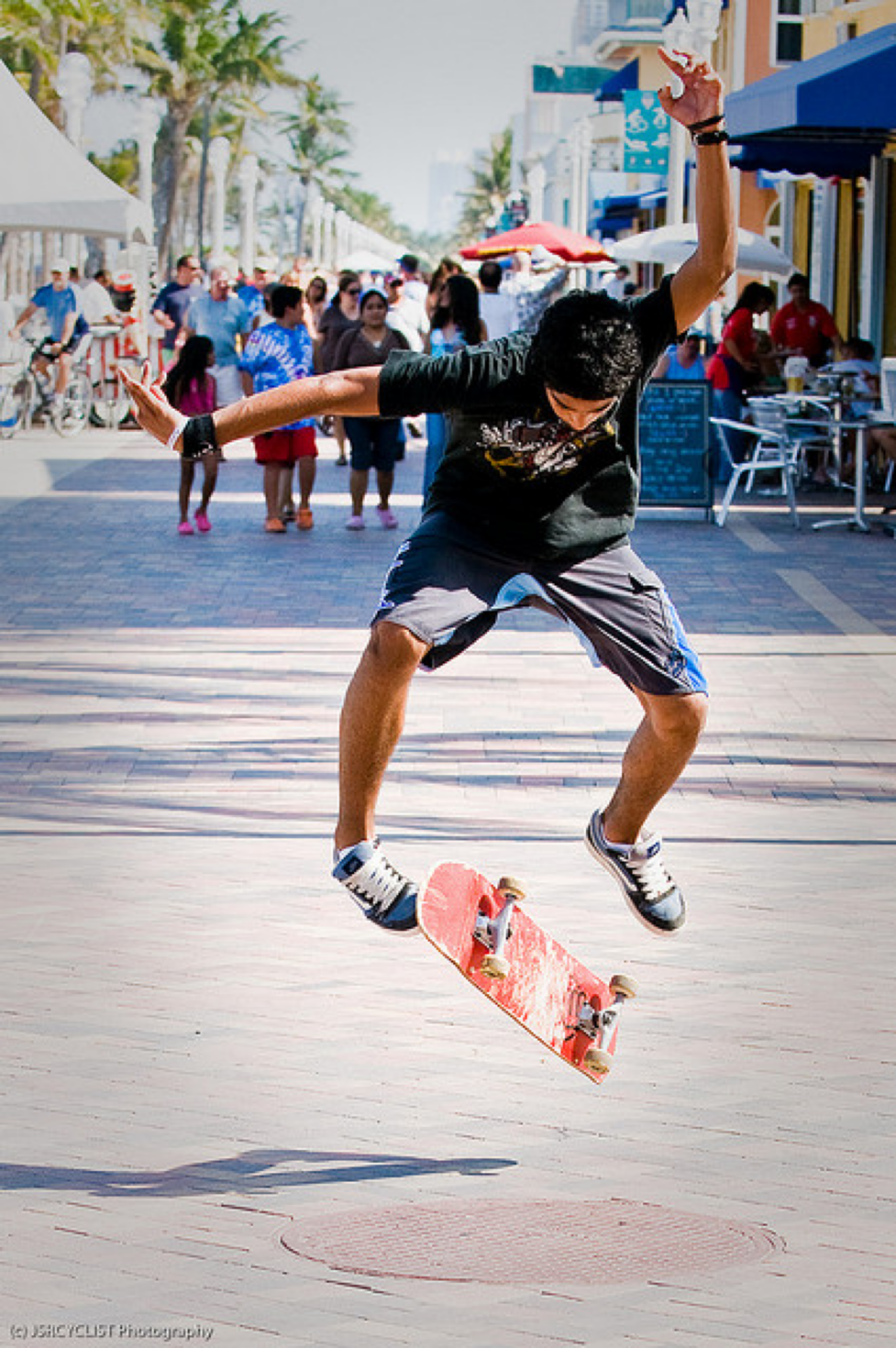}
        \caption{Original: Person}
        \label{fig:threshold}
    \end{subfigure}%
    \begin{subfigure}[t]{\divCoco\textwidth}
        \centering
        \includegraphics[height=\hsb]{177-skateboard-COCO_val2014_000000087038-orig.png}
        \caption{Original: Skateboard}
        \label{fig:threshold}
    \end{subfigure}%
    
    %%%%%%%%%%%%%%%%%%%%%%%%%%%%%%%%%%%%%%%%%%%%%%%%%%%%%%%%%%%%%%%%%%%%%%%%%%%%%%%%%%%%%%%%%%%%%%%%%%%%%
    \begin{subfigure}[t]{\divCoco\textwidth}
        \centering
        \includegraphics[height=\hsheep]{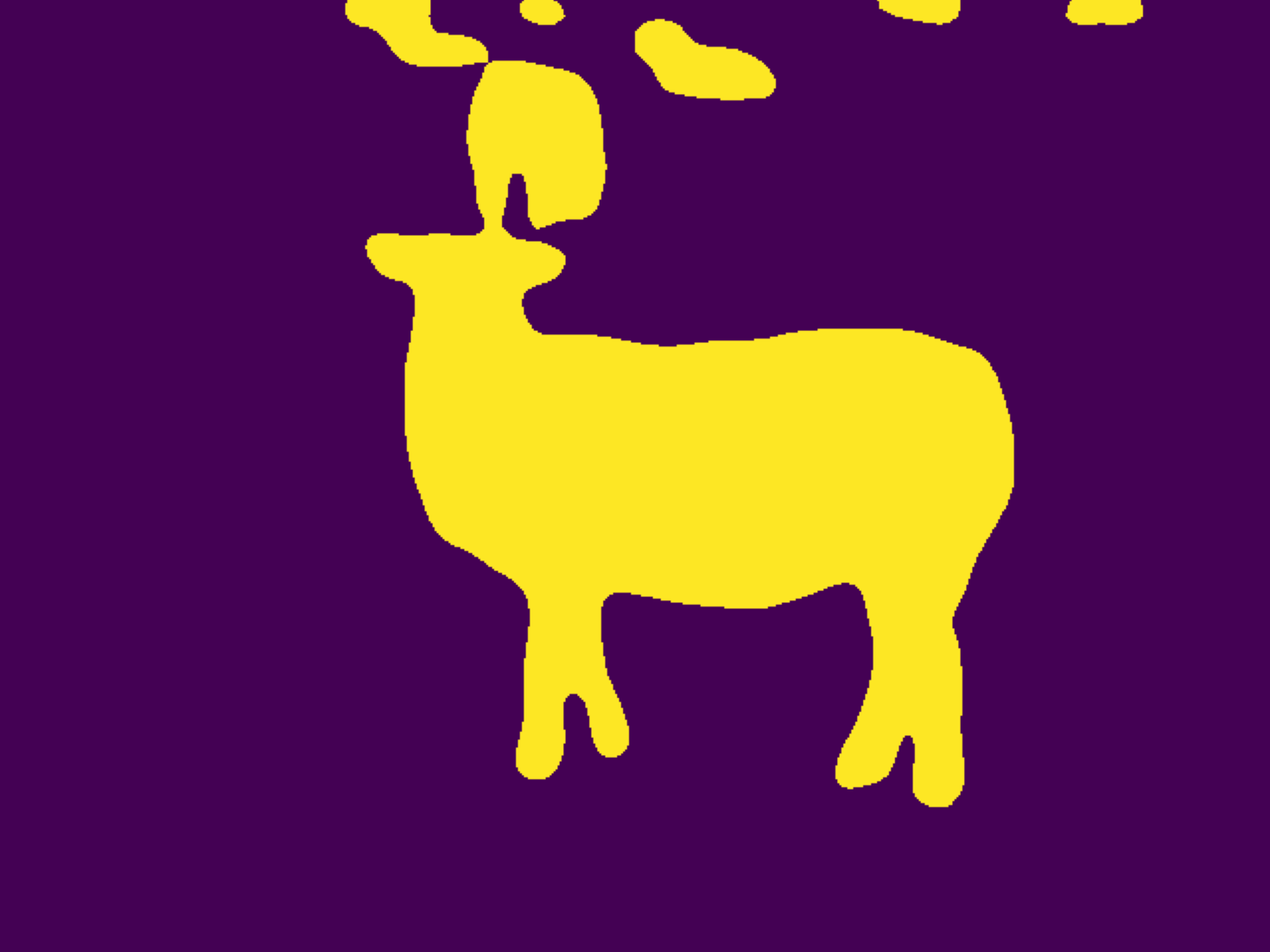}
        \caption{Ours: Sheep}
        \label{fig:threshold}
    \end{subfigure}%
    \begin{subfigure}[t]{\divCoco\textwidth}
        \centering
        \includegraphics[height=\hfridge]{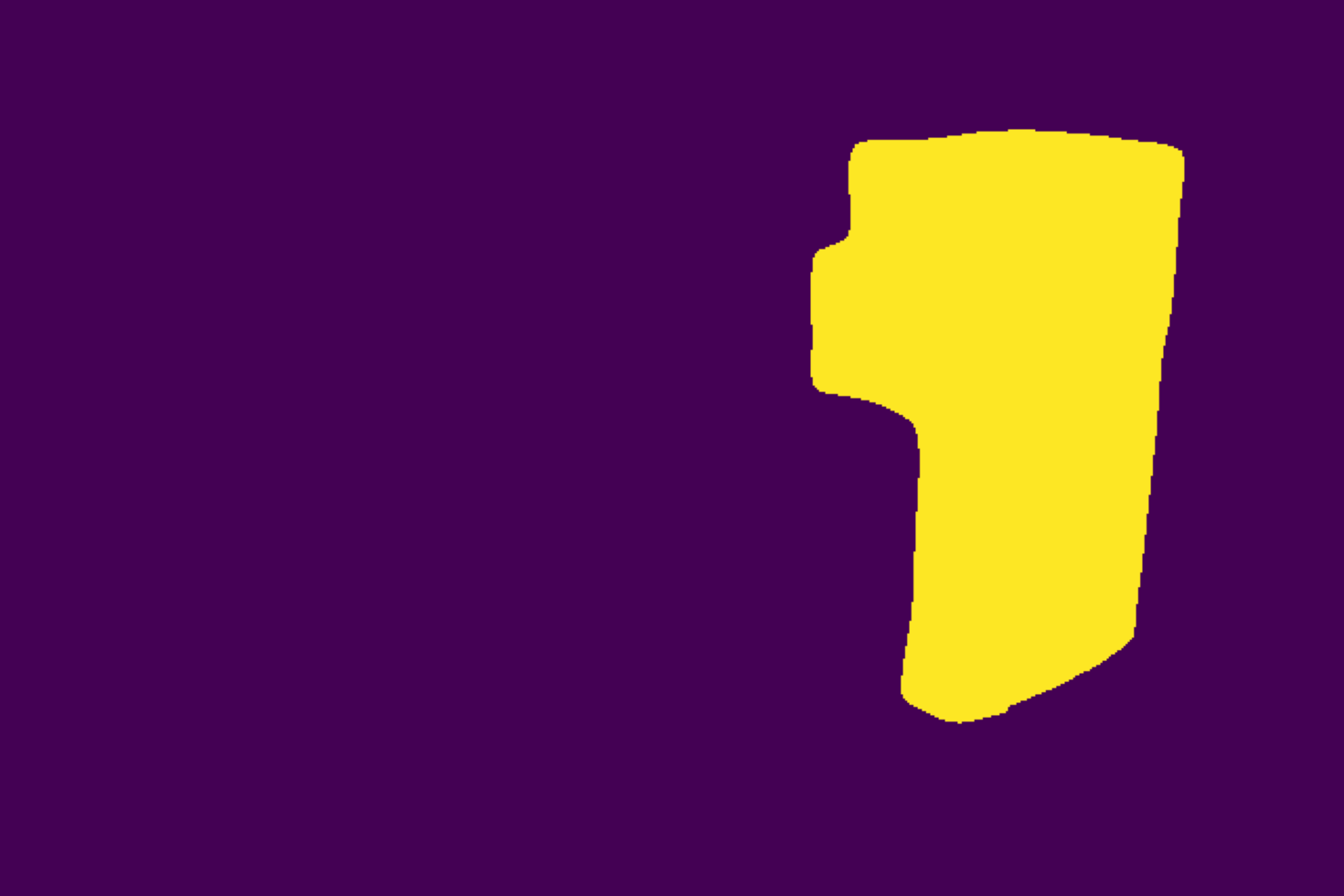}
        \caption{Ours: Refrigerator}
        \label{fig:threshold}
    \end{subfigure}%
    \begin{subfigure}[t]{\divCoco\textwidth}
        \centering
        \includegraphics[height=\hperson]{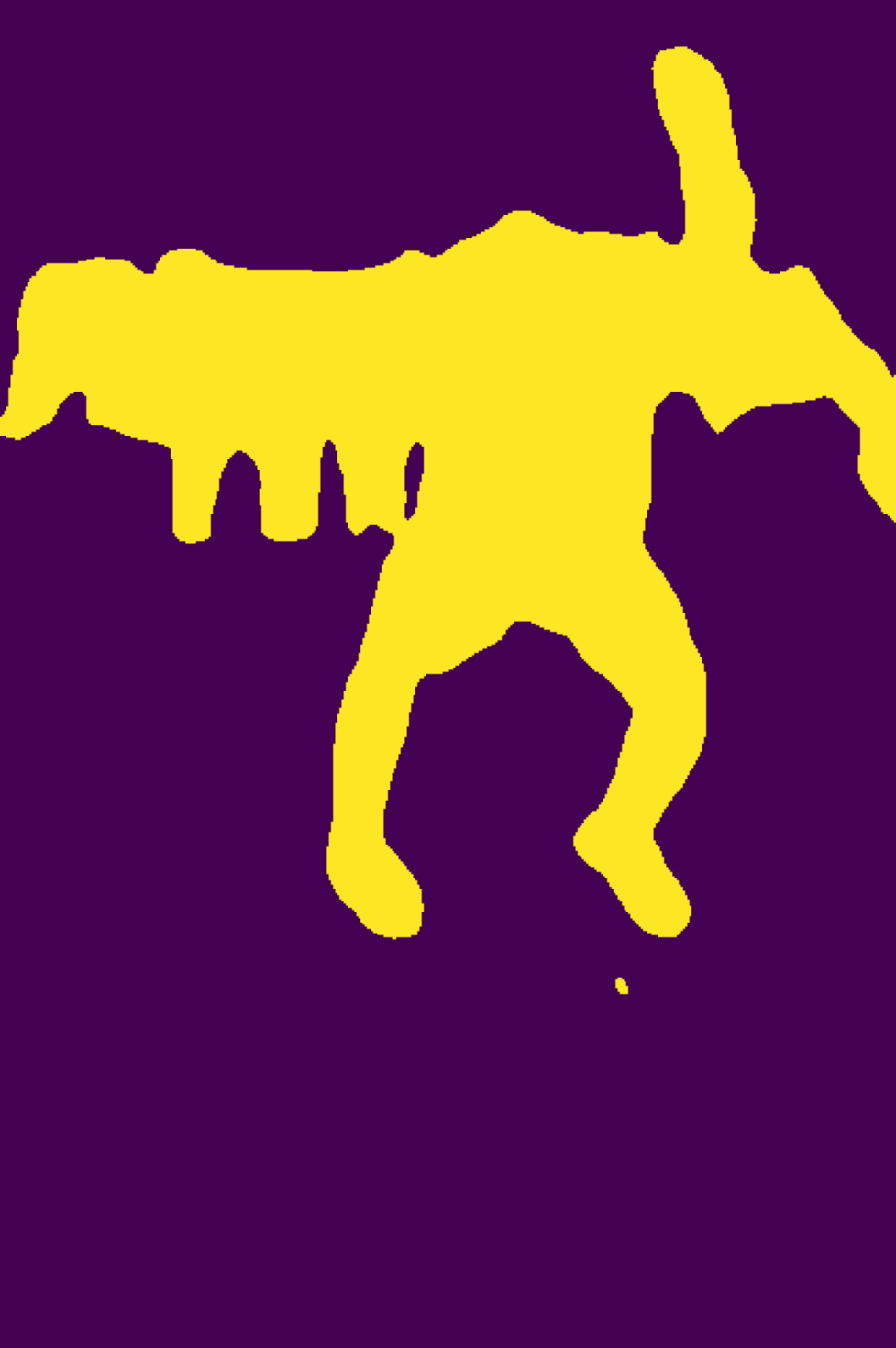}
        \caption{Ours: Person}
        \label{fig:threshold}
    \end{subfigure}%
    \begin{subfigure}[t]{\divCoco\textwidth}
        \centering
        \includegraphics[height=\hsb]{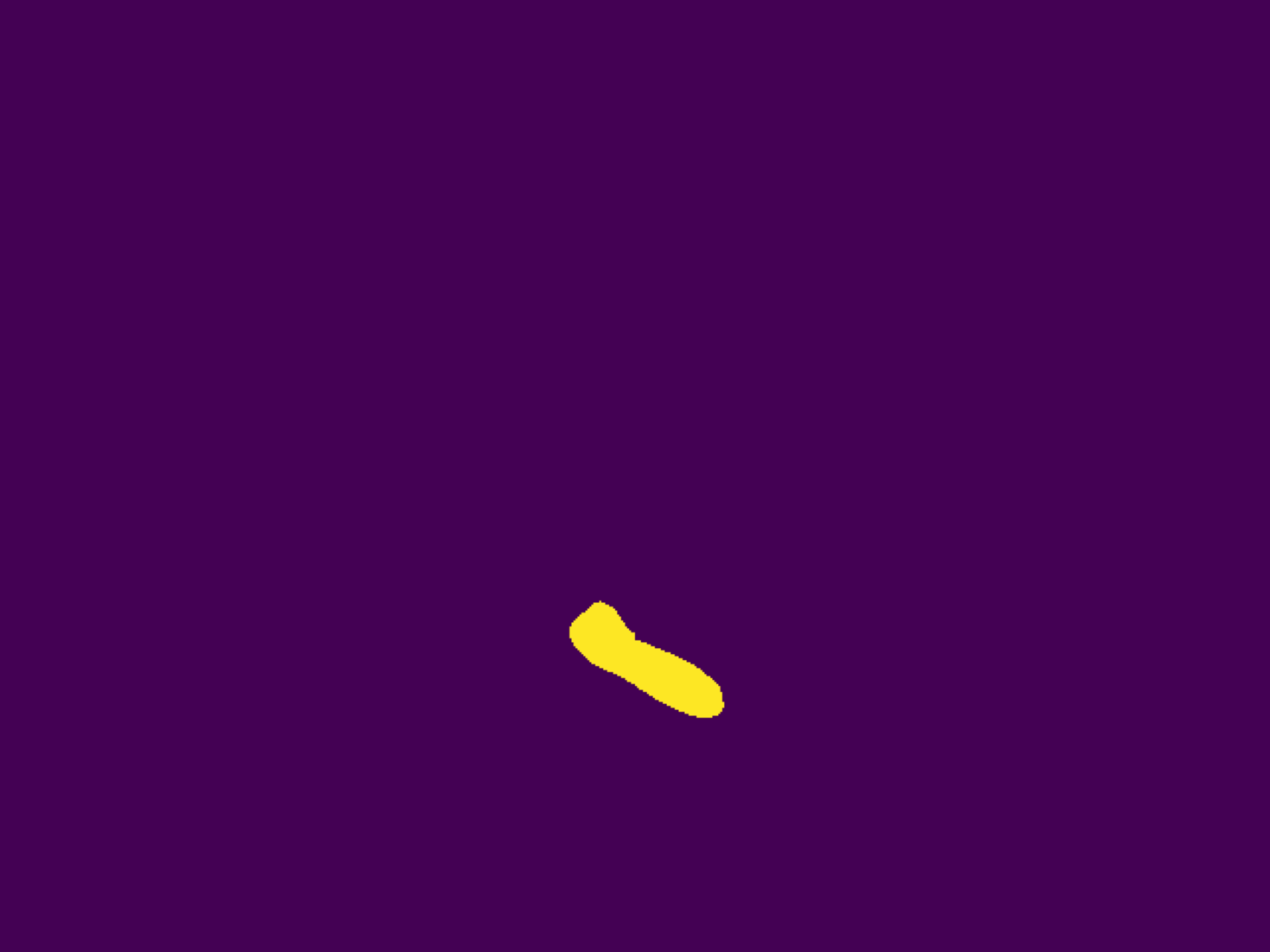}
        \caption{Ours: Skateboard}
        \label{fig:threshold}
    \end{subfigure}%
    
    %%%%%%%%%%%%%%%%%%%%%%%%%%%%%%%%%%%%%%%%%%%%%%%%%%%%%%%%%%%%%%%%%%%%%%%%%%%%%%%%%%%%%%%%%%%%%%%%%%%%%
    \begin{subfigure}[t]{\divCoco\textwidth}
        \centering
        \includegraphics[height=\hsheep]{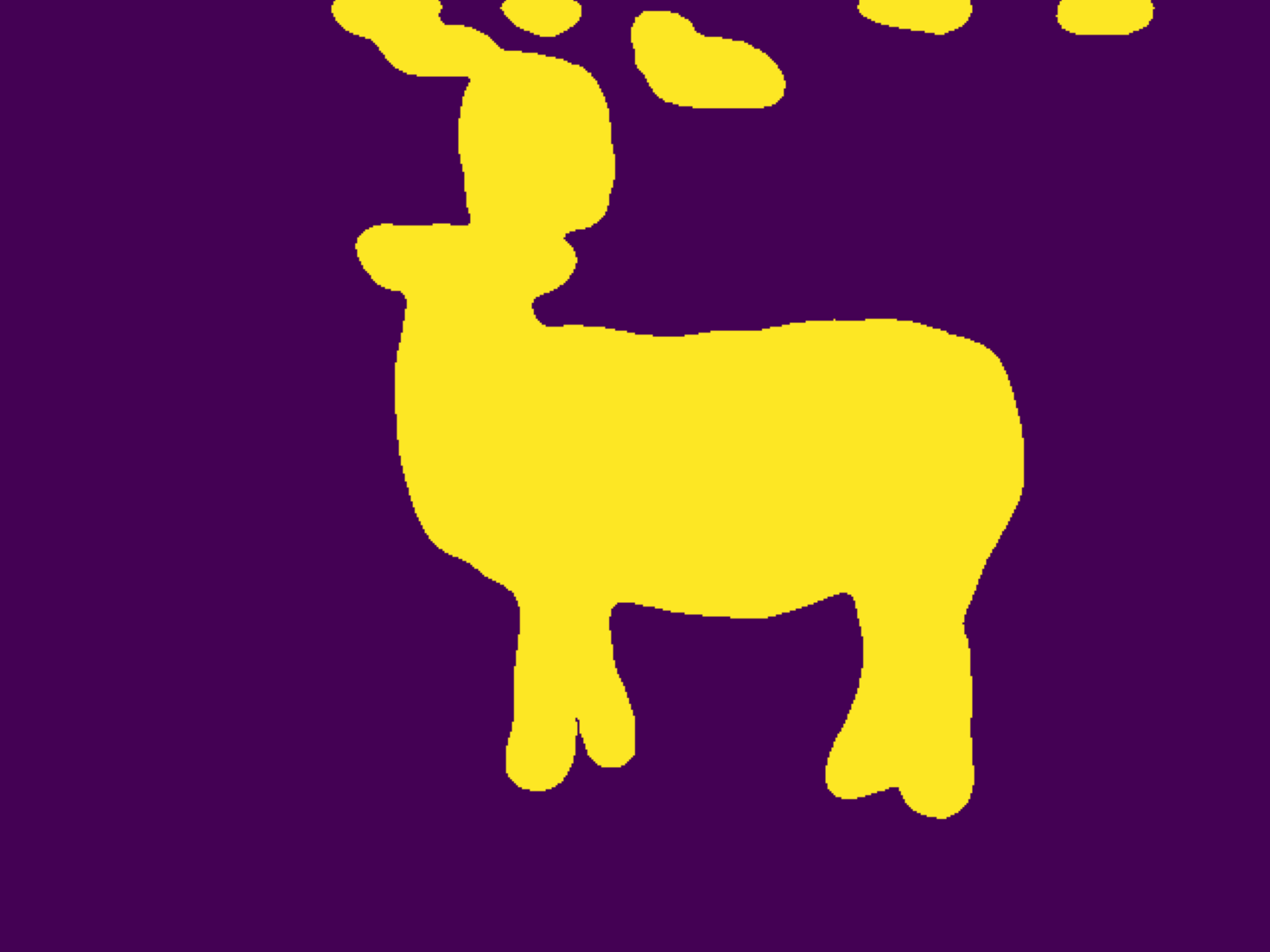}
        \caption{Baseline: Sheep}
        \label{fig:method}
    \end{subfigure}%
    \begin{subfigure}[t]{\divCoco\textwidth}
        \centering
        \includegraphics[height=\hfridge]{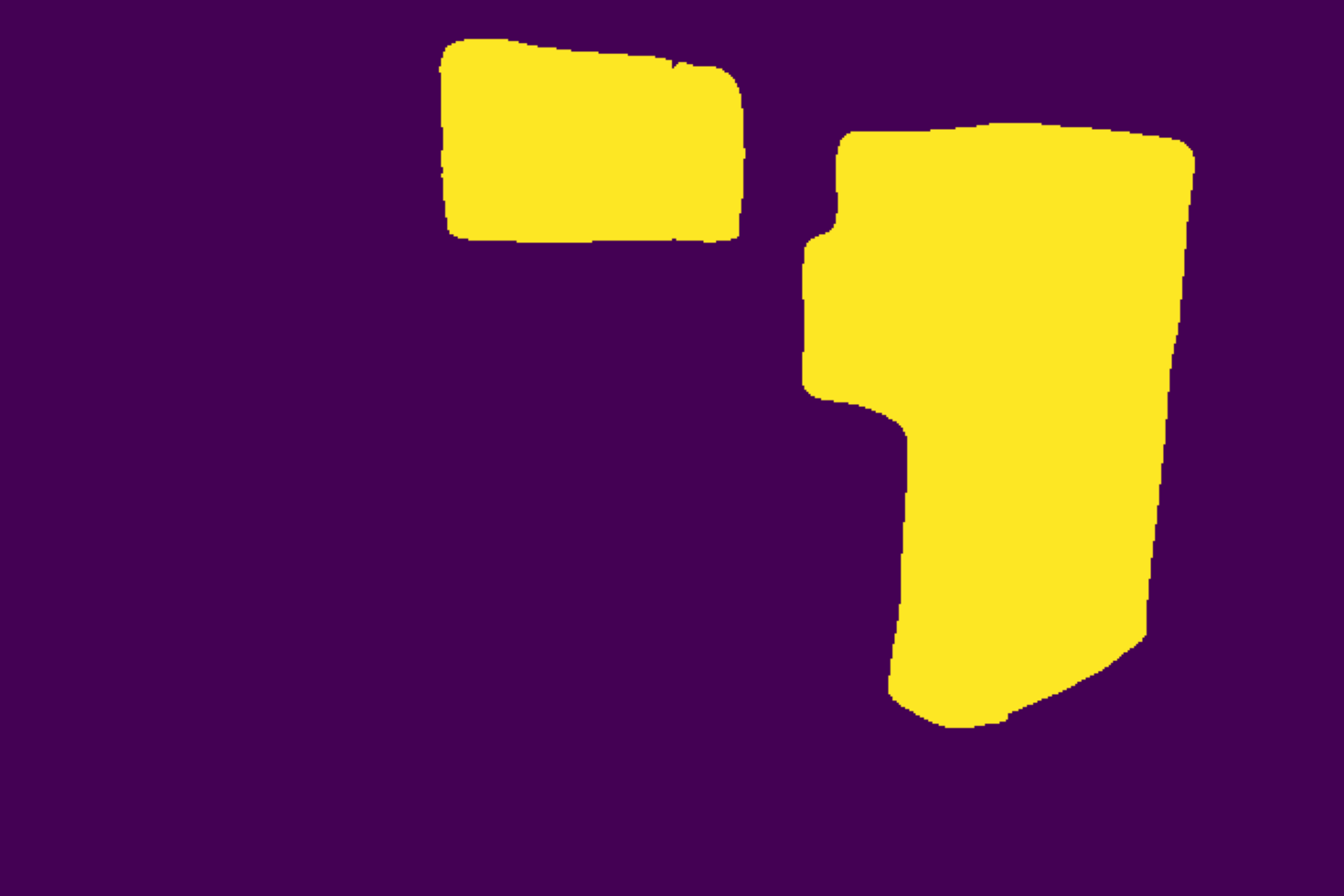}
        \caption{Baseline: Refrigerator}
        \label{fig:method}
    \end{subfigure}%
    \begin{subfigure}[t]{\divCoco\textwidth}
        \centering
        \includegraphics[height=\hperson]{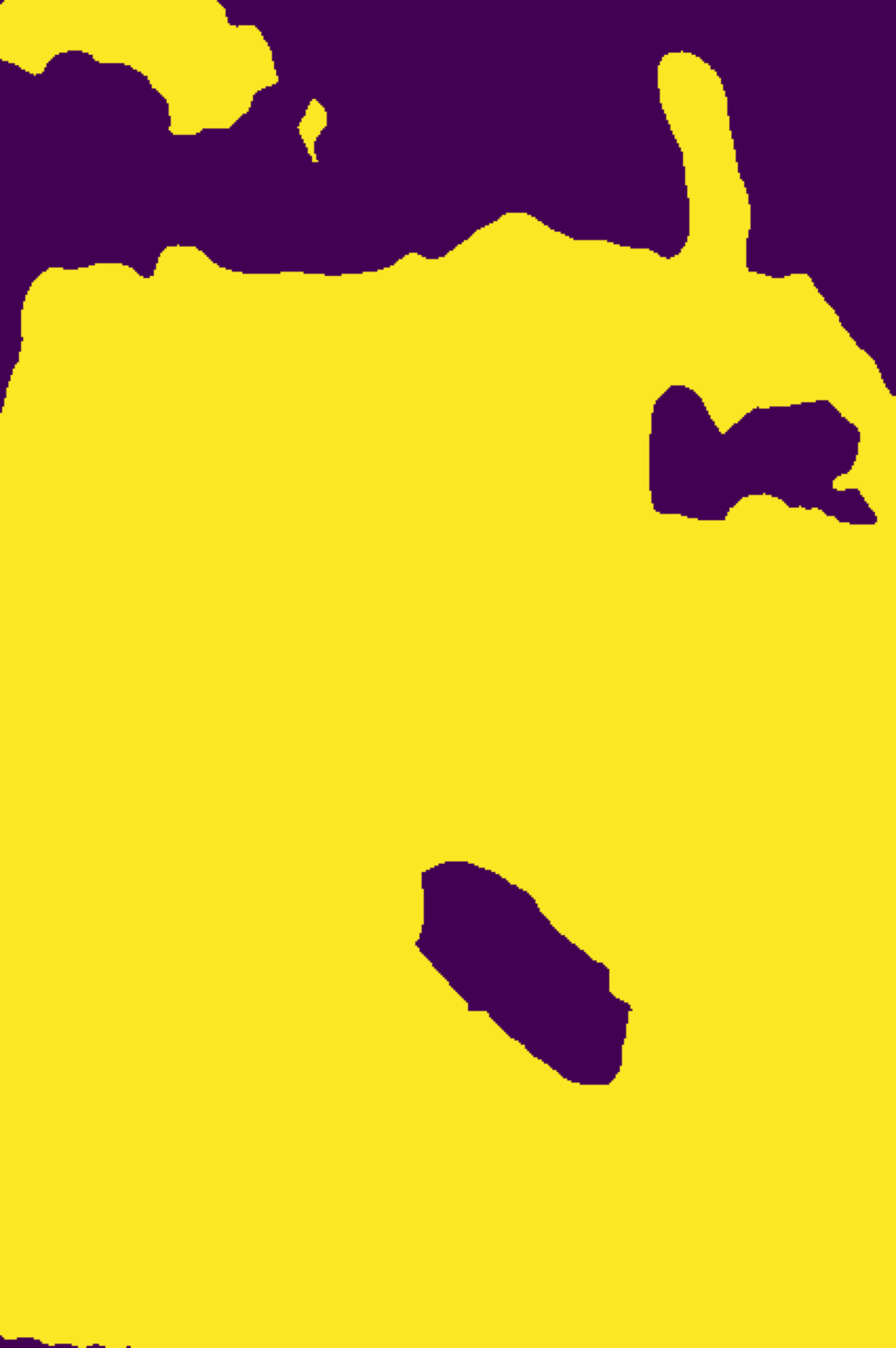}
        \caption{Baseline: Person}
        \label{fig:method}
    \end{subfigure}%
    \begin{subfigure}[t]{\divCoco\textwidth}
        \centering
        \includegraphics[height=\hsb]{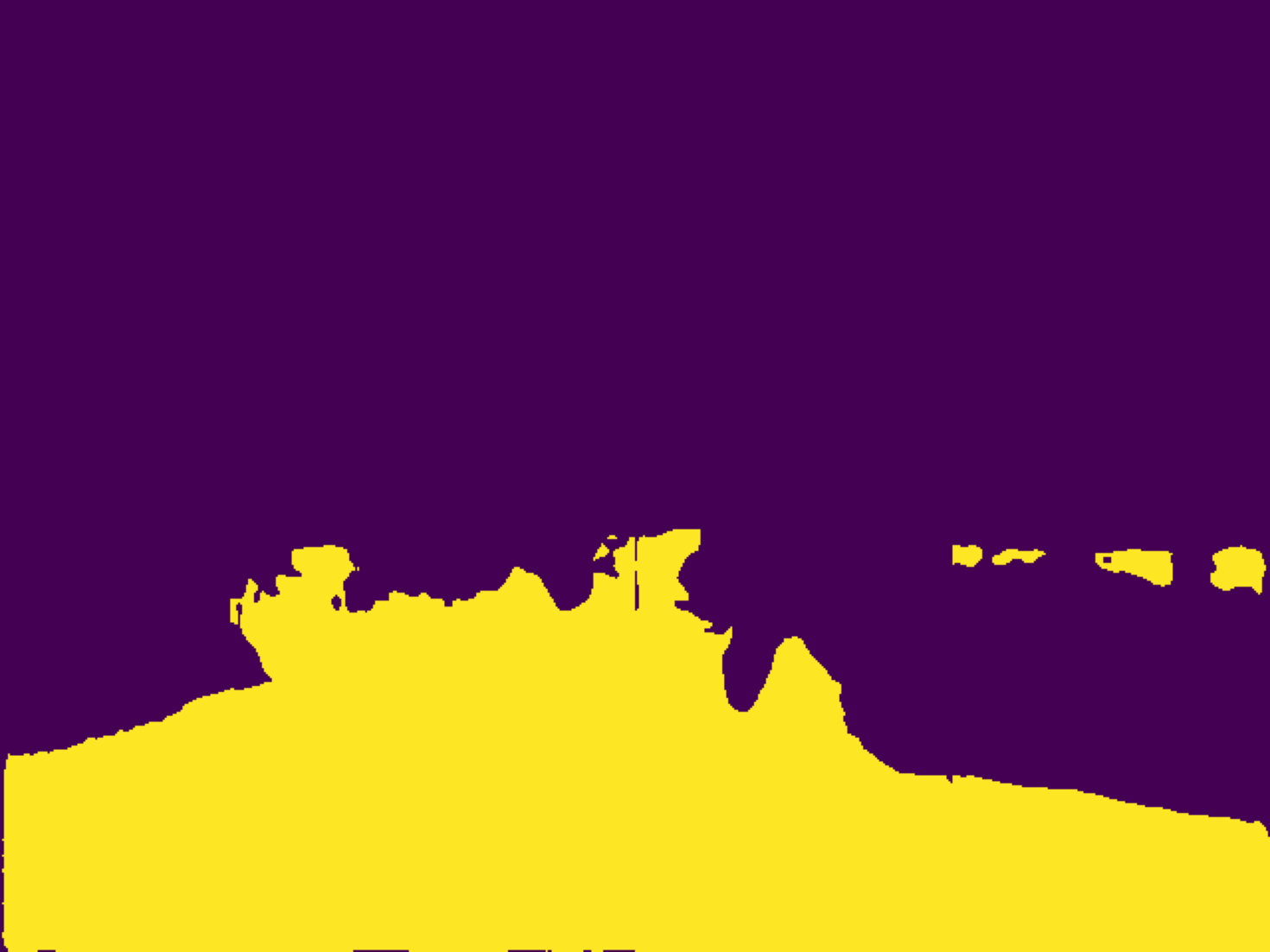}
        \caption{Baseline: Skateboard}
        \label{fig:method}
    \end{subfigure}%
    \caption{ 
        Examples of predictions from the COCO data set. The original image is shown on the top row, our method using an SVM classifier is on the second row, and the baseline is on the third row. Our method is generally slightly more accurate in contouring objects, as shown in the first column, and can distinguish related objects better, shown in the second column, and is less prone to overprediction, as shown in the third and fourth columns.
    }
    \label{fig:coco}
    \figvspace{}
\end{figure*}

%
%
%
%
%
%%%%%%%%%%%%%%%%%%%%%%%%%%%%%%%%%%%%%%%%%%%%%%%%%%%%%%%%%%%%%%%%%%%%%%%%%%%%%%%%%%%%%%%%%%%%%%%%%%%%%
\subsection{Image benchmarks}
\label{sec:results-img}

From the results of the image benchmark experiments, presented in Table \ref{tab:image-benchmarks}, it can be seen that our method yields increases of $15\%$ in F1-score on COCO, $9\%$ in Pascal Context 459, and $15\%$ in the Matterport data set. This supports the hypothesis that the knowledge of the distribution in language space is beneficial for visual classification. 

Notably, recall is higher in the baseline. This is likely because the method of comparing the query with a single antonym splits the embedding space in two by a hyperplane. This leads to half of the embedding space being predicted as positive, which can encompass the correct region in its entirety; however, it conversely yields a high number of false positives, as reflected in the low precision. Our method can alleviate this problem by providing a more accurate estimate of the correct volume in the embedding space.

Examples of predictions from the COCO data set are shown in Figure \ref{fig:coco}. While most of the predictions are very similar, our method performs more accurately in contouring the objects, as shown in the first column. Another common occurrence is that the baseline misclassifies visually similar and semantically close objects, such as a refrigerator and a microwave oven. In contrast, our method can distinguish between these concepts, as shown in the second column. At times, the single complement is so far from the image embeddings that the baseline dramatically extends the prediction to unrelated areas, shown in the third and fourth columns.

%
%
%
%
%
%%%%%%%%%%%%%%%%%%%%%%%%%%%%%%%%%%%%%%%%%%%%%%%%%%%%%%%%%%%%%%%%%%%%%%%%%%%%%%%%%%%%%%%%%%%%%%%%%%%%%
\subsection{Querying maps}
\label{sec:results-maps}

From the results of the map querying experiment, presented in Table \ref{tab:map-experiment}, it can be seen that, with LSeg, our method using ground truth synonyms cannot significantly improve over the baseline, and our method with \gls{llm}-generated synonyms performs worse than the baseline. This highlights the sensitivity of the choice of the synonyms and antonyms. Still, ultimately, this result is congruent with the hypothesis that with a good approximation of $\mathcal{Q}$, the queryability results do improve.

Example queries from the data set are shown in Figure \ref{fig:queries-img}, where predicted binary masks from our method with the ground truth synonyms and antonyms, along with the baseline, are shown. From the Figure, we can see that the differences in LSeg are very minor. However, with OpenSeg, the baseline exhibits a high number of false positives, reflected in the low precision, exemplified in Figure \ref{fig:queries-img}f and \ref{fig:queries-img}g, whereas our method finds the correct voxels from the map in corresponding Figures \ref{fig:queries-img}b and \ref{fig:queries-img}c. The precision-recall curve of our method could be further optimized, as our method can sometimes suffer from reduced recall, as shown in the example of a failure case in Figure \ref{fig:queries-img}d.

\newcommand\imageScaleThree{0.1}
\newcommand\divThree{0.25}

\begin{figure*}[ht!]
    \begin{subfigure}[t]{\divThree\textwidth}
        \centering
        \scalebox{\imageScaleThree}{\includegraphics{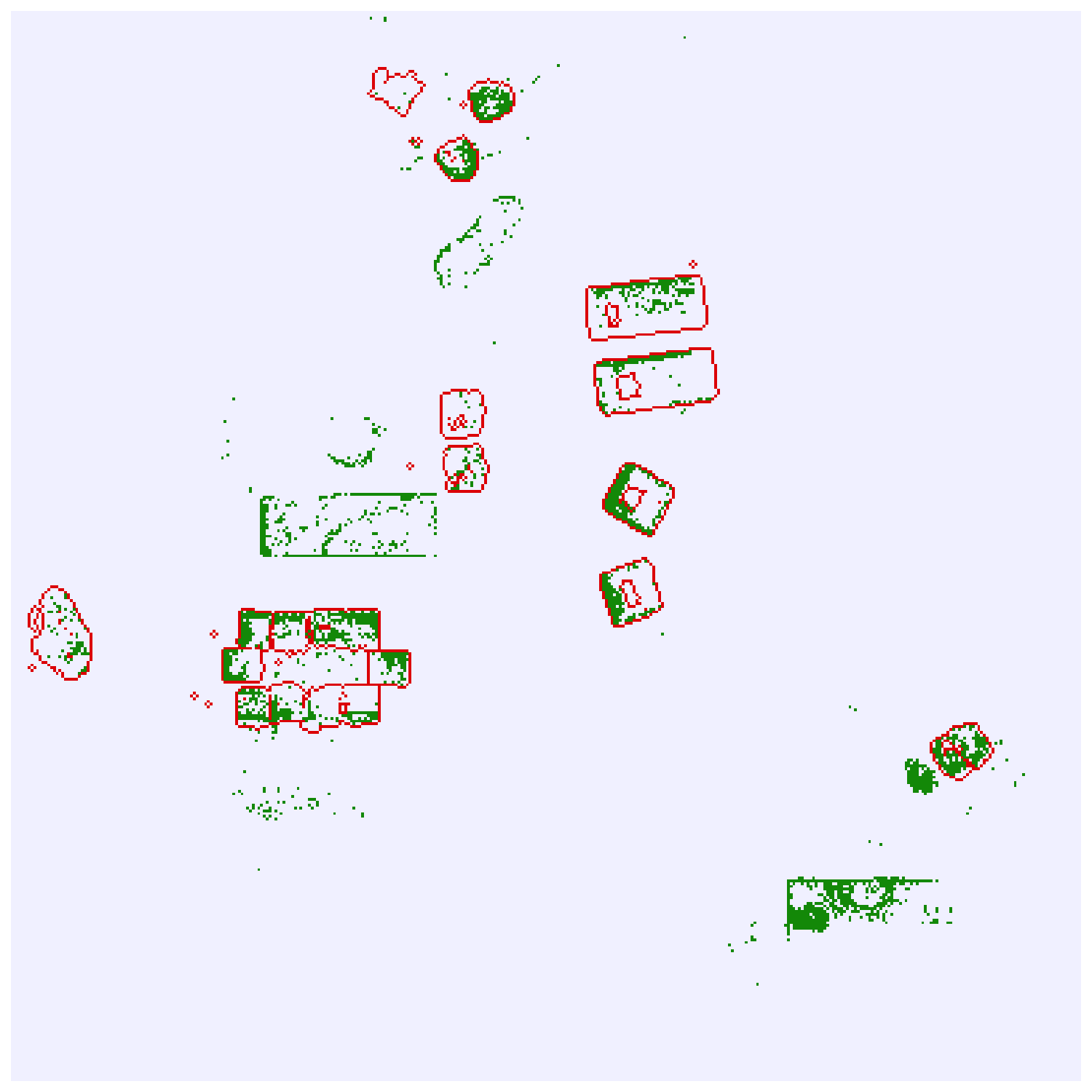}}
        \caption{Ours, LSeg: Chair}
        \label{fig:threshold}
    \end{subfigure}%
    \begin{subfigure}[t]{\divThree\textwidth}
        \centering
        \scalebox{\imageScaleThree}{\includegraphics{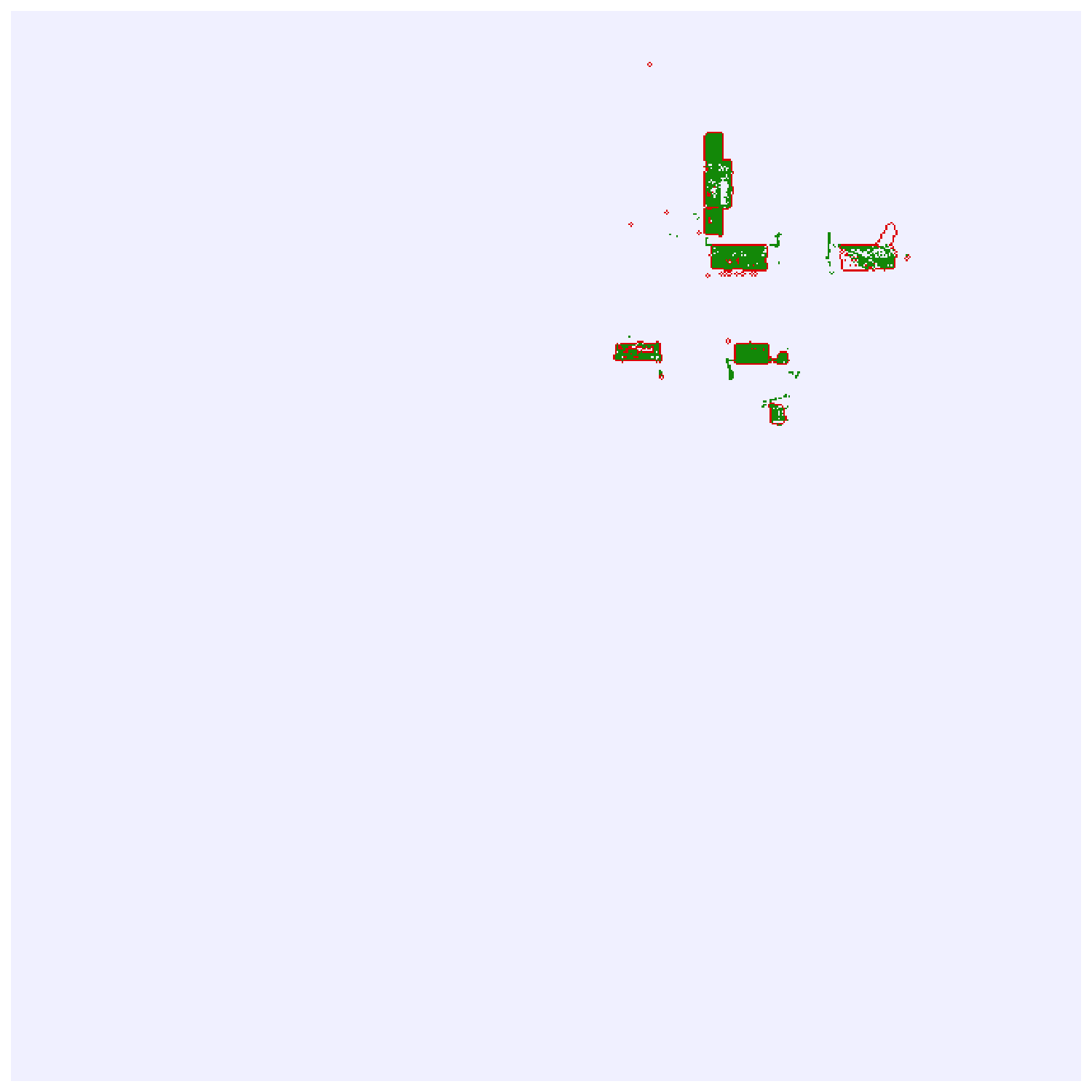}}
        \caption{Ours, OpenSeg: Shelving}
        \label{fig:threshold}
    \end{subfigure}%
    \begin{subfigure}[t]{\divThree\textwidth}
        \centering
        \scalebox{\imageScaleThree}{\includegraphics{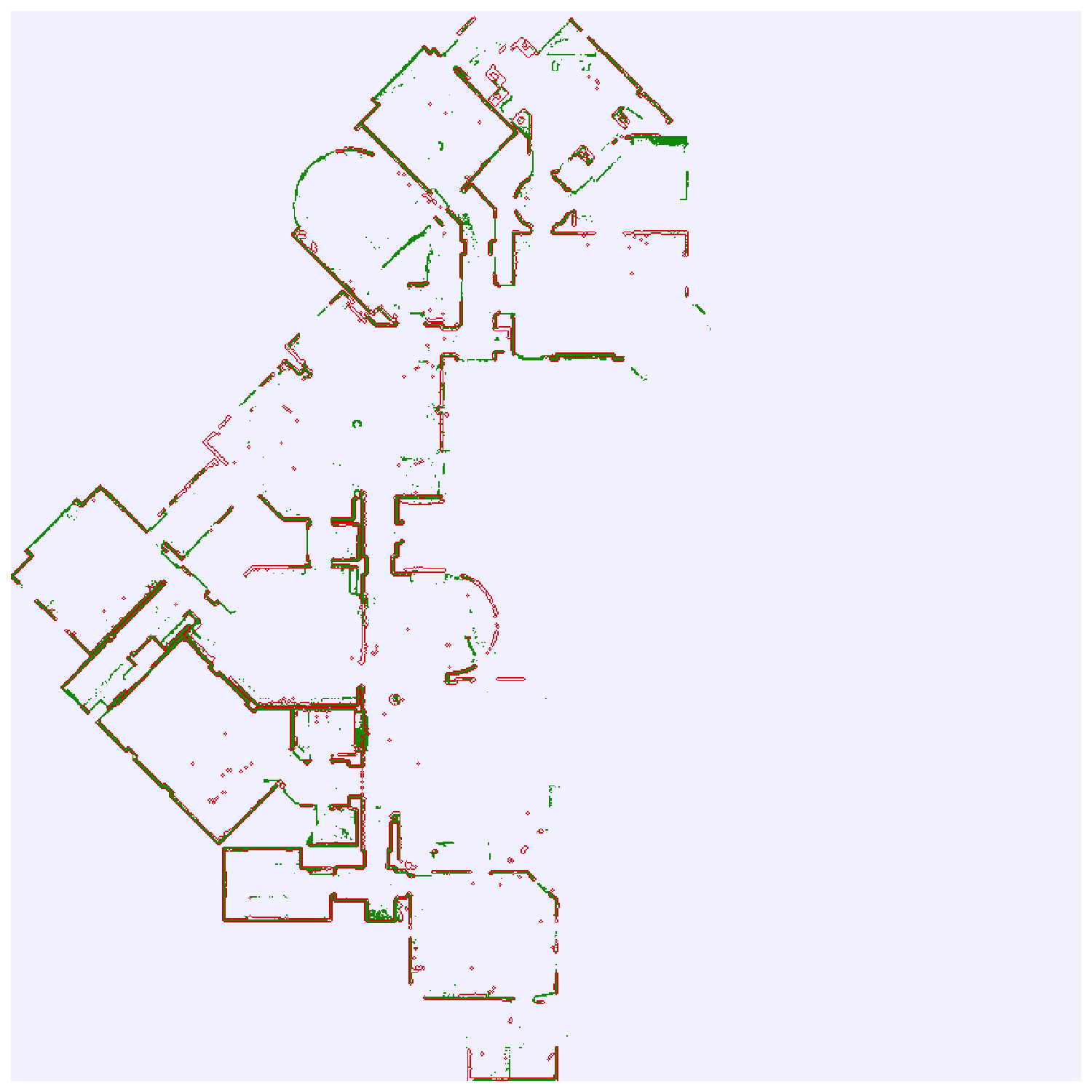}}
        \caption{Ours, OpenSeg: Wall}
        \label{fig:threshold}
    \end{subfigure}%
    \begin{subfigure}[t]{\divThree\textwidth}
        \centering
        \scalebox{\imageScaleThree}{\includegraphics{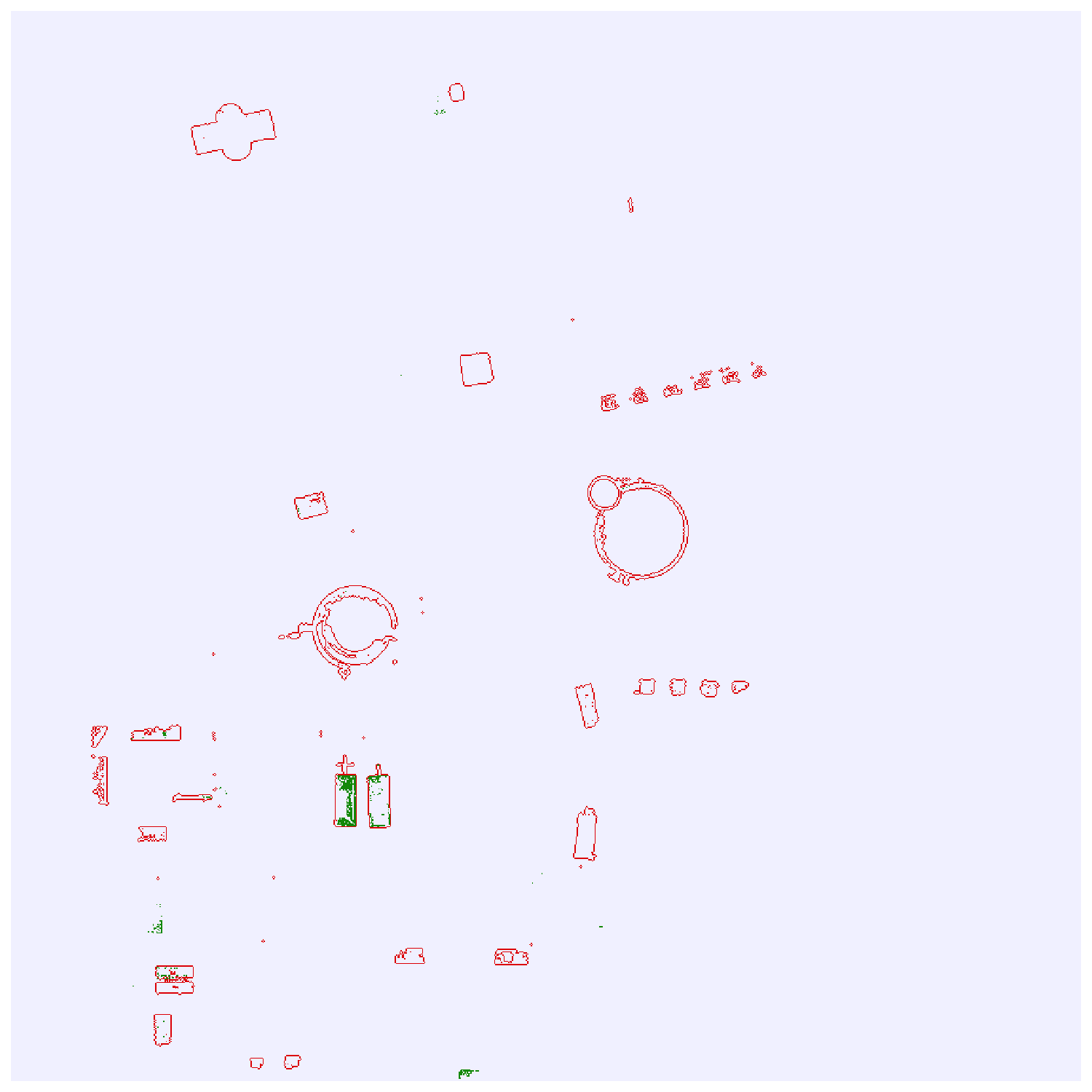}}
        \caption{Ours, OpenSeg: Table}
        \label{fig:threshold}
    \end{subfigure}%

    \begin{subfigure}[t]{\divThree\textwidth}
        \centering
        \scalebox{\imageScaleThree}{\includegraphics{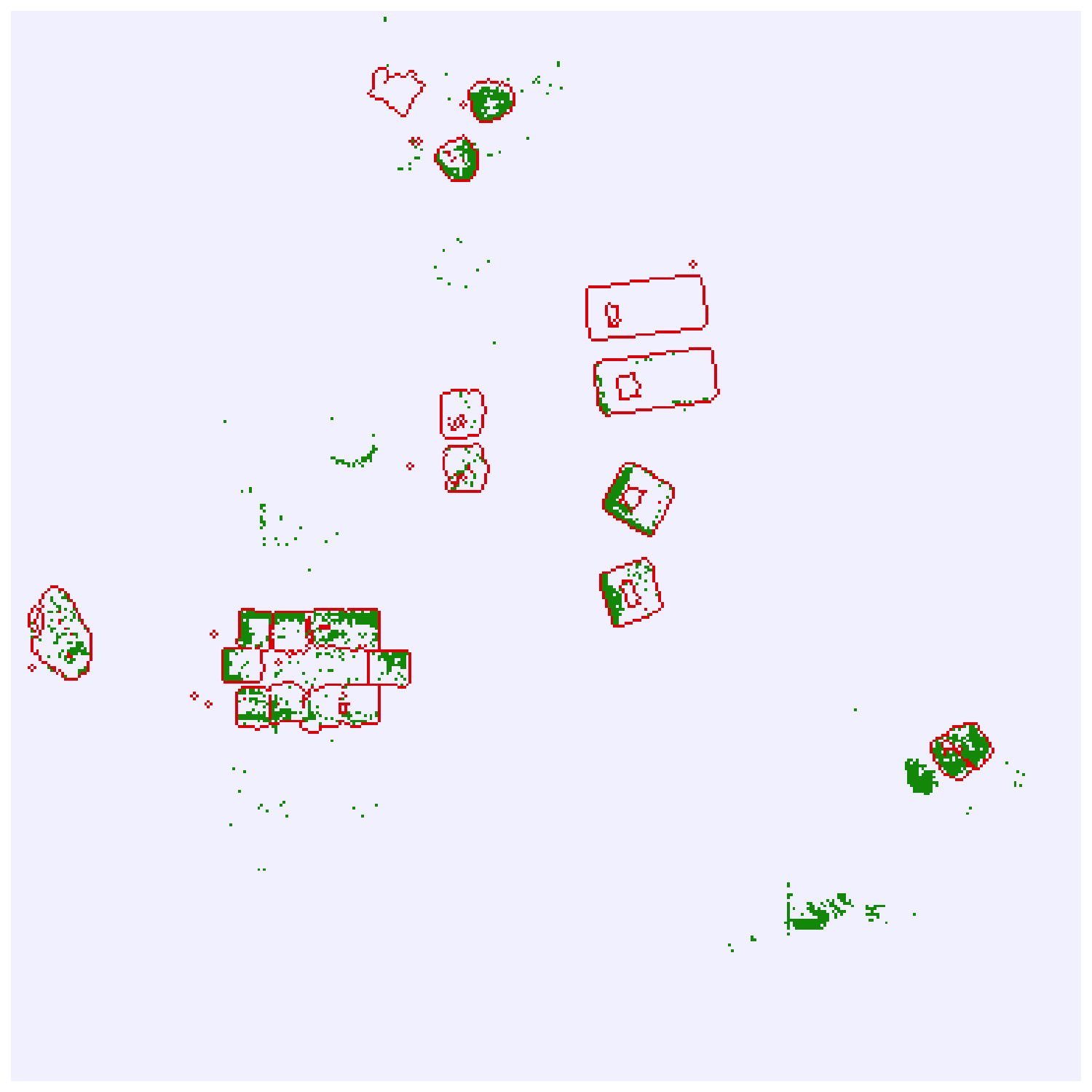}}
        \caption{Baseline, LSeg: Chair}
        \label{fig:method}
    \end{subfigure}%
    \begin{subfigure}[t]{\divThree\textwidth}
        \centering
        \scalebox{\imageScaleThree}{\includegraphics{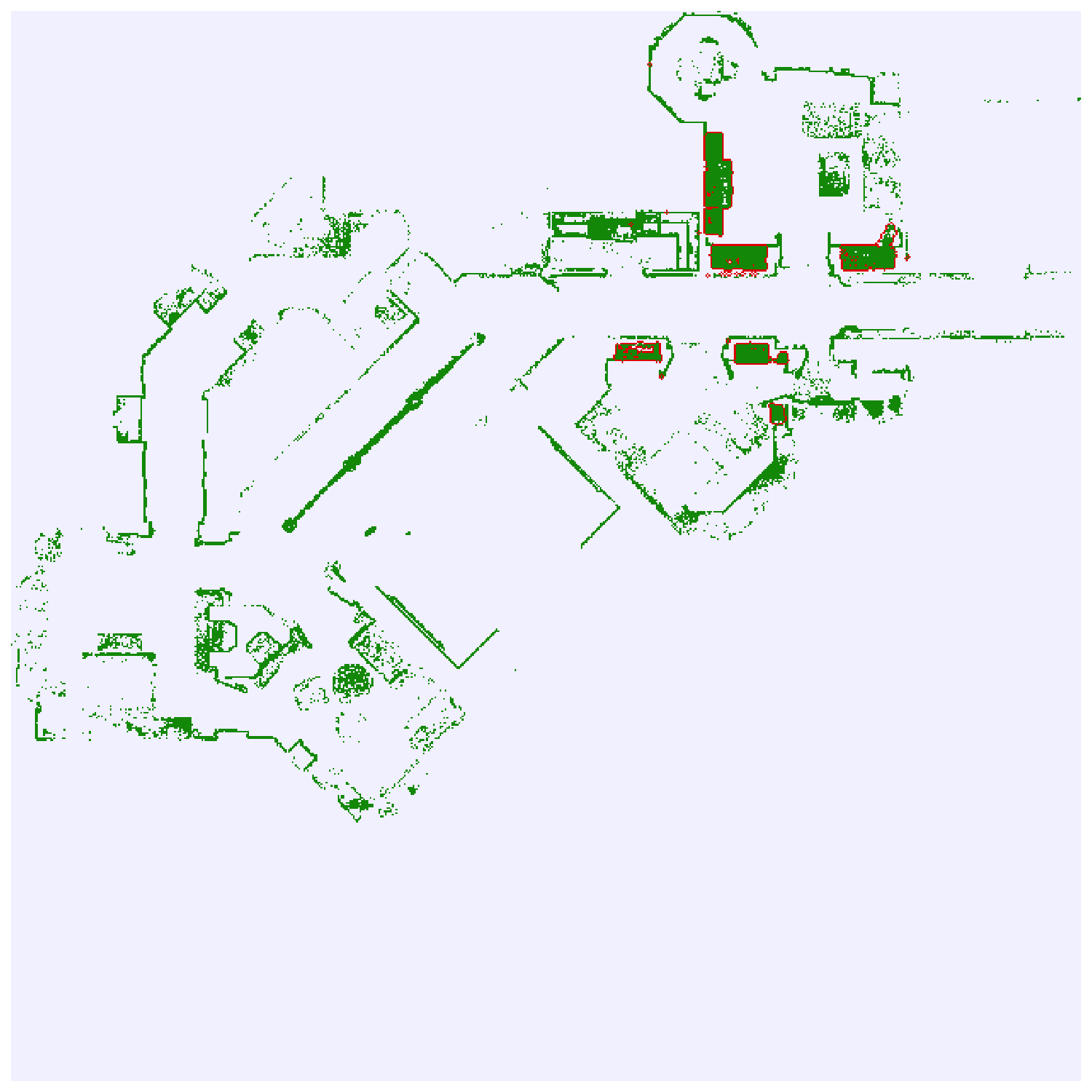}}
        \caption{Baseline, OpenSeg: Shelving}
        \label{fig:method}
    \end{subfigure}%
    \begin{subfigure}[t]{\divThree\textwidth}
        \centering
        \scalebox{\imageScaleThree}{\includegraphics{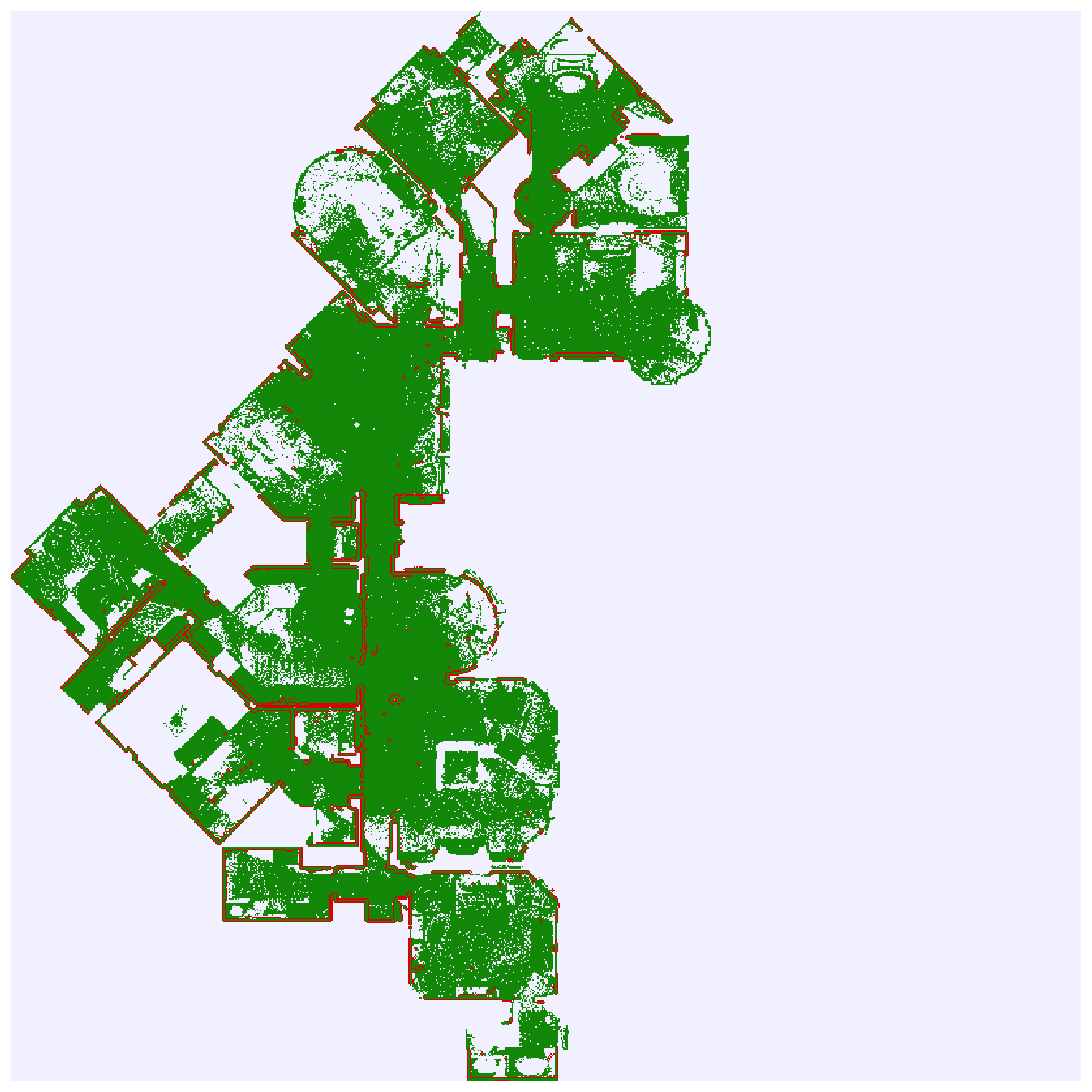}}
        \caption{Baseline, OpenSeg: Wall}
        \label{fig:method}
    \end{subfigure}%
    \begin{subfigure}[t]{\divThree\textwidth}
        \centering
        \scalebox{\imageScaleThree}{\includegraphics{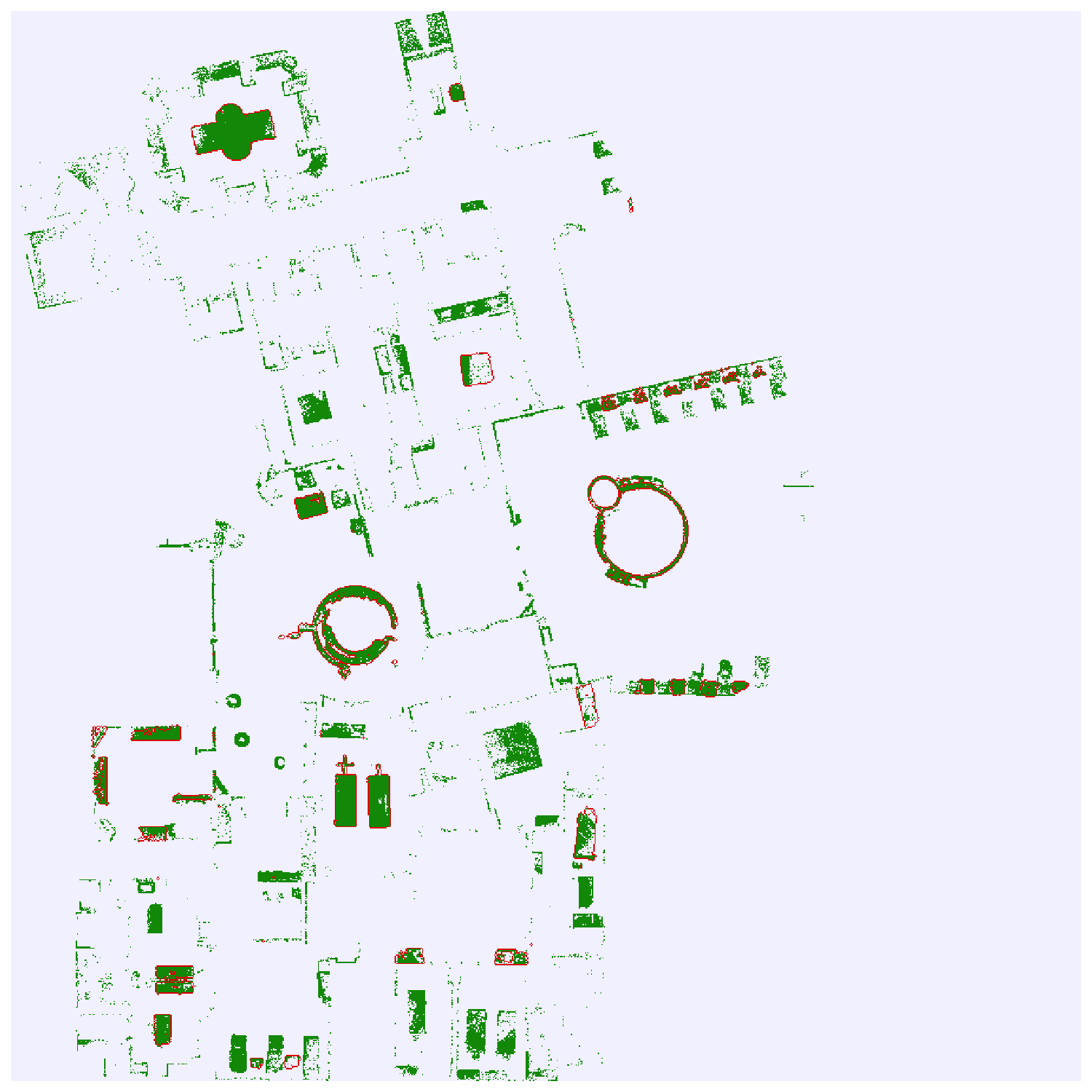}}
        \caption{Baseline, OpenSeg: Table}
        \label{fig:method}
    \end{subfigure}%
    \caption{ 
        Examples of map query experiment results. Our method with \qt{ground truth} synonyms is shown in the top row, and the baseline is shown in the bottom row, with the same query displayed in each column. With LSeg (a), (e), the results are very similar. On the other hand, with OpenSeg, we can clearly see the difference, particularly in the significantly improved precision of our method. However, the drawback is that our method suffers from reduced recall, as evident in the failure cases in (d) and (h). The predictions are shown in green, and the outline of the ground truth in red.
    }
    \label{fig:queries-img}
    \figvspace{}
\end{figure*}

When using OpenSeg, our method outperforms the baseline with $227\%$ increase in F1-score. Even with \gls{llm}-generated synonyms and antonyms, our method yields $78\%$ improvement on the F1-score. This significant increase suggests the potential for using encoders other than LSeg in robotic applications, where LSeg has performed the best with existing querying methods so far. 

While OpenSeg appears to be more sensitive to thresholding, leading to lower overall results, the vocabulary seems to be broader \cite{pekkanen_2025_latent_semantics}. Although OpenSeg results are overall worse than LSeg, the results indicate that our method is effective at classifying embeddings from different encoders.

Interestingly, the results differ significantly between the Matterport data set and the maps. This implies that a modality gap exists between the maps and the images. The maps aggregate the measurements originating from the image data set, so the aggregation is likely the leading cause of this modality gap. 

In VLMaps, the environment is represented as a voxel map, and therefore multiple embeddings are aggregated in each voxel. The mean of the embeddings is used as the aggregate. This can cause problems, especially in cases where the embeddings are multi-modal within the cell. Typically, each mode results from a set of measurements in which an object with a specific semantics is identified. However, the aggregate embedding produced by the mean of the distribution may not accurately reflect the original semantics, but may be a new concept altogether. 

We evaluated different aggregation methods for the voxels, including the geometric median, selecting a representative sample, and training a transformer encoder to aggregate the embeddings. However, the geometric median yielded only $4\%$ increase in F1-score, and therefore, cannot address this problem.

%--%--%--%--%--%--%--%--%--%--%--%--%--%--%--%--%--%--%--%--%--%--%--%--%--%--%--%--
\begin{table}[t]
    \caption{Comparison of map querying methods in maps}
    \tabcapspace{}
    \begin{center}
        \setlength\tabcolsep{6pt}
        \scalebox{1.0}{
                \begin{tabular}{llllll}
                    \toprule
                    \textbf{Encoder} & \textbf{Method} & \textbf{F1} & \textbf{IoU} & \textbf{Precision} & \textbf{Recall} \\
                    \midrule
                    \multirow{3}{*}{LSeg}       & baseline         & 0.561          & 0.395          & 0.536          & 0.589          \\
                                                & our ground truth & \textbf{0.574} & \textbf{0.408} & \textbf{0.567} & 0.583          \\
                                                \cmidrule{2-6}
                                                & our LLM          & 0.396          & 0.252          & 0.289          & \textbf{0.656} \\
                    \midrule
                    \multirow{3}{*}{OpenSeg}    & baseline         & 0.136          & 0.073          & 0.073          & \textbf{0.926} \\
                                                & our ground truth & \textbf{0.445} & \textbf{0.296} & \textbf{0.622} & 0.369          \\
                                                \cmidrule{2-6}
                                                & our LLM          & 0.242          & 0.139          & 0.154          & 0.586          \\
                    \bottomrule
                \end{tabular}
        }
        \setlength\tabcolsep{6pt}
        \label{tab:map-experiment}
    \end{center}
    \figvspace{}
\end{table}

%--%--%--%--%--%--%--%--%--%--%--%--%--%--%--%--%--%--%--%--%--%--%--%--%--%--%--%--

%
%
%
%
%
%%%%%%%%%%%%%%%%%%%%%%%%%%%%%%%%%%%%%%%%%%%%%%%%%%%%%%%%%%%%%%%%%%%%%%%%%%%%%%%%%%%%%%%%%%%%%%%%%%%%%
\subsection{Ablation studies}
\label{sec:results-ablation}

\subsubsection{Prompt engineering}
The ablation results from prompt engineering are presented in Table \ref{tab:ablation-prompt}. The prompt engineering yields benefits only on the baseline with LSeg, as it was proposed for that purpose. 

Interestingly, prompt engineering does not yield benefits with OpenSeg, whereas in the original work, prompt engineering was evaluated, but only in conjunction with ensembling, making a direct comparison difficult. This suggests that not all heuristics used for improving queries are directly generalizable across encoders. With our method, the prompt engineering did not yield improved performance. 

This is understandable, as our method already uses more embeddings to estimate the distribution $\mathcal{Q}$. From this experiment, we can see that the majority of the gap between the baseline and our method is due to the prompt engineering. Prompt engineering aims to generalize the specific query to capture better the region $\mathcal{Q}$. 

Even if ultimately only one embedding is used, the mean of the 63 prompts is better located in the embedding space to capture $\mathcal{R}_{\mathcal{Q}}$, and therefore this result is congruent to our hypothesis.

%--%--%--%--%--%--%--%--%--%--%--%--%--%--%--%--%--%--%--%--%--%--%--%--%--%--%--%--
\begin{table}[t]
    \caption{The effect of prompt engineering (PE) in map querying}
    \tabcapspace{}
    \begin{center}
        \setlength\tabcolsep{6pt}
        \scalebox{1.0}{
                \begin{tabular}{llllll}
                    \toprule
                    \textbf{Method} & \textbf{Encoder} & \textbf{PE} & \textbf{F1} & \textbf{Precision} & \textbf{Recall} \\
                    \midrule
                    \multirow{2}{*}{baseline} & \multirow{2}{*}{LSeg}    & yes & \textbf{0.561} & \textbf{0.536} & 0.589          \\
                                              &                          & no  & 0.407          & 0.301          & \textbf{0.651} \\
                    \midrule
                    \multirow{2}{*}{baseline} & \multirow{2}{*}{OpenSeg} & yes & 0.136          & 0.073          & 0.926          \\
                                              &                          & no  & 0.136          & 0.074          & 0.931          \\
                    \midrule
                    \multirow{2}{*}{C-SVM}    & \multirow{2}{*}{LSeg}    & yes & 0.396          & 0.289          & 0.656          \\ 
                                              &                          & no  & 0.396          & 0.289          & 0.656          \\
                    \midrule
                    \multirow{2}{*}{C-SVM}    & \multirow{2}{*}{OpenSeg} & yes & 0.242          & 0.154          & 0.586          \\     
                                              &                          & no  & 0.242          & 0.154          & 0.586          \\
                    \bottomrule
                \end{tabular}
        }
        \setlength\tabcolsep{6pt}
        \label{tab:ablation-prompt}
    \end{center}
    \figvspace{}
\end{table}

%--%--%--%--%--%--%--%--%--%--%--%--%--%--%--%--%--%--%--%--%--%--%--%--%--%--%--%--

%--%--%--%--%--%--%--%--%--%--%--%--%--%--%--%--%--%--%--%--%--%--%--%--%--%--%--%--

\renewcommand{\thefootnote}{\fnsymbol{footnote}}
\begin{table}[t]
    \caption{The comparison of antonyms in map querying with baseline}
    \tabcapspace{}
    \centering
    \begin{threeparttable}
    \setlength\tabcolsep{6pt}
    \begin{tabular}{lllll}
        \toprule
        \textbf{Encoder} & \textbf{Antonym} & \textbf{F1} & \textbf{Precision} & \textbf{Recall} \\
        \midrule
        \multirow{9}{*}{LSeg}    & other                & \textbf{0.561} & \textbf{0.536} & 0.589          \\
                                 & others               & 0.546          & 0.500          & 0.604          \\
                                 & background           & 0.463          & 0.363          & 0.649          \\
                                 & texture              & 0.460          & 0.361          & 0.657          \\
                                 & stuff                & 0.451          & 0.346          & 0.651          \\
                                 & nothing              & 0.393          & 0.279          & 0.680          \\
                                 & object               & 0.358          & 0.240          & 0.709          \\
                                 & things               & 0.346          & 0.231          & 0.698          \\
                                 & zero vector\tnote{*} & 0.048          & 0.025          & \textbf{1.000} \\
        \midrule
        \multirow{9}{*}{OpenSeg} & other                & 0.136          & 0.073          & 0.926          \\
                                 & others               & 0.127          & 0.068          & 0.936          \\
                                 & background           & 0.131          & 0.071          & 0.935          \\
                                 & texture              & \textbf{0.193} & \textbf{0.110} & 0.870          \\
                                 & stuff                & 0.114          & 0.061          & 0.947          \\
                                 & nothing              & 0.185          & 0.104          & 0.890          \\
                                 & object               & 0.118          & 0.063          & 0.934          \\
                                 & things               & 0.123          & 0.066          & 0.939          \\
                                 & zero vector\tnote{*} & 0.065          & 0.033          & \textbf{0.991} \\
        \bottomrule
    \end{tabular}
    \begin{tablenotes}
        \item[*] Prompt engineering not used
    \end{tablenotes}
    \setlength\tabcolsep{6pt}
    \figvspace{}
    \end{threeparttable}
    \label{tab:ablation-antonym}
\end{table}
\renewcommand{\thefootnote}{\arabic{footnote}}
%--%--%--%--%--%--%--%--%--%--%--%--%--%--%--%--%--%--%--%--%--%--%--%--%--%--%--%--

%--%--%--%--%--%--%--%--%--%--%--%--%--%--%--%--%--%--%--%--%--%--%--%--%--%--%--%--
\begin{table}[t]
    \caption{The comparison of different query set generation methods in map querying with our method with LSeg}
    \tabcapspace{}
    \begin{center}
        \setlength\tabcolsep{6pt}
        \scalebox{1.0}{
                \begin{tabular}{lllll}
                    \toprule
                    \textbf{Synonyms} & \textbf{Antonyms} & \textbf{F1} & \textbf{Precision} & \textbf{Recall} \\
                    \midrule
                    ground truth & ground truth & \textbf{0.576} & 0.565          & \textbf{0.587} \\
                    adjectives   & ground truth & \textbf{0.575} & \textbf{0.583} & 0.568          \\
                    adjectives   & LLM          & 0.555          & 0.534          & 0.581          \\
                    ground truth & LLM          & 0.528          & 0.469          & 0.608          \\
                    adjectives   & lexicon      & 0.518          & 0.438          & 0.639          \\
                    LLM          & ground truth & 0.462          & 0.476          & 0.469          \\
                    LLM          & LLM          & 0.396          & 0.289          & 0.656          \\
                    ground truth & lexicon      & 0.378          & 0.261          & 0.697          \\
                    adjectives   & adjectives   & 0.362          & 0.246          & 0.700          \\
                    ground truth & adjectives   & 0.279          & 0.174          & 0.740          \\
                    LLM          & lexicon      & 0.236          & 0.141          & 0.729          \\
                    LLM          & adjectives   & 0.142          & 0.078          & 0.859          \\
                    lexicon      & adjectives   & 0.049          & 0.025          & 0.995          \\
                    lexicon      & lexicon      & 0.048          & 0.025          & 0.994          \\
                    lexicon      & ground truth & 0.040          & 0.036          & 0.047          \\
                    lexicon      & LLM          & 0.024          & 0.013          & 0.116          \\
                    \bottomrule
                \end{tabular}
        }
        \setlength\tabcolsep{6pt}
        \label{tab:ablation-synonyms}
    \end{center}
    \figvspace{}
\end{table}

%--%--%--%--%--%--%--%--%--%--%--%--%--%--%--%--%--%--%--%--%--%--%--%--%--%--%--%--

\subsubsection{Antonyms}
The results of the ablation experiment with different antonyms are presented in Table \ref{tab:ablation-antonym}. With LSeg, \qt{other} is the best single word. This is expected, as it is used in LSeg as the background label, and subsequently proposed to be used in VLMaps. 

Interestingly, the distribution of the results differs between LSeg and OpenSeg; in LSeg, there is a greater spread between any two words, where the interval is $0.215$, whereas in OpenSeg, the interval is only $0.079$, without taking into account the zero vectors, which perform significantly worse than any other antonym with both encoders.

\subsubsection{Synonyms}
The results of the ablation experiment in terms of synonym generation are presented in Table \ref{tab:ablation-synonyms}. A suitable set of synonyms is essential to match the targets accurately, as evident from the low precision of the lexicon, which serves as the random baseline. However, to get the best results, a suitable antonym set is needed in conjunction with the synonyms. This is highlighted, \eg{} with the adjective antonyms reducing the accuracy of the ground truth synonyms.

The ground truth provides the best overall results, as none of the synonyms are other classes present in the evaluation dataset, and antonyms correspond to the other classes present in the environment, which effectively enhances the precision of the detections. These results are the best overall query results in maps and support the hypothesis that knowledge of language distribution improves visual classification results.

While the adjective method performs well, it cannot be used in querying specific object instances. In this experiment, we evaluate the queryability of generic classes (\eg{} \qt{chair}, \qt{table}, \qt{painting}), which are more effectively captured by adjectives that expand the region in the embedding space. However, when a particular object is queried (\eg{} \qt{the red metal chair}), the adjectives force the generalization of the query, therefore making it unusable in an open-vocabulary context. However, adjectives perform poorly as antonyms, as they are in the form adjective + \qt{other}, which is semantically meaningless.

The LLM is the most versatile method, as it can produce synonyms for any word. The downside is the unreliability of the predictions as well as the presence of other classes in the synonyms, which reduces the prediction accuracy.

\section{Conclusion}
\label{sec:conclusion}

In this work, we argue that using a single query embedding and a single antonym is not sufficient to perform visual classification on \gls{vlm} embeddings in maps or images. Leveraging the semantic knowledge to heuristically sample from the language space improves visual classification results due to the shared latent visual-language embedding space of \glspl{vlm}. We show this with the proposed method, where the cosine similarity metric is replaced by a classifier, yielding improved results in image data sets and maps when using OpenSeg. 

Notably, our method yields a significant improvement when utilizing OpenSeg as the encoder. This opens the possibility of using other encoders besides the widely adopted LSeg. While the embedding space of OpenSeg is more sensitive to thresholding, the latent space seems to have a more expansive vocabulary, making it more suitable for open-vocabulary querying.

In our ablation experiments, we explore a range of commonly used heuristics for querying embeddings. We verify that prompt engineering phrases have to be designed for a particular encoder. We examine the various negative examples used in the literature and observe different behaviors between the tested encoders. This leads to the conclusion that these heuristics do not readily generalize beyond their designed environments.

% important
These results are applicable in a wide range of robotic applications, as our methods are not dependent on a particular encoder or even a modality of the visual space, whether maps or images. Many of the recent methods proposed for robotic object retrieval today threshold the cosine similarity metric or contrast the query with a single negative example. Future work could be devoted to extending these methods by replacing thresholding with a classifier, possibly yielding significantly improved results, while retaining their other distinctive benefits.

However, several open questions remain, particularly in the open-vocabulary setting, where it remains open how to find the best possible query set. In this work, we evaluated the queryability of generic classes, but how can the queryability be evaluated in an open-vocabulary setting? Furthermore, as the majority of the novel computer vision methods have originally been proposed to work in the image domain, how to bridge the modality gap between images and maps? These questions need answers to bring robotic mapping into the era of open vocabulary semantics.

\bibliographystyle{IEEEtran}
\bibliography{consolidated-cleaned-abbr,IEEEabrv,ctrl}
\end{document}